\documentclass[lettersize,journal]{IEEEtran}
\usepackage{amsmath,amsfonts}
\usepackage{algorithmic}
\usepackage{algorithm}
\usepackage{array}
\usepackage[caption=false,font=scriptsize]{subfig}
\usepackage{textcomp}
\usepackage{stfloats}
\usepackage{url}
\usepackage{verbatim}
\usepackage{graphicx}
\usepackage{cite}
\usepackage{multirow}
\usepackage{makecell}
\usepackage{tabularx} 
\usepackage{xcolor}
\usepackage{booktabs}
\usepackage{graphicx}
\usepackage{xcolor}
\usepackage{diagbox}
\usepackage{textcomp}
\usepackage[table,xcdraw]{xcolor}
\usepackage{amsthm}

\begin{document}

\title{BadSKP: Backdoor Attacks on Knowledge Graph-Enhanced LLMs with Soft Prompts}

\author{Xiaoting Lyu, Yufei Han, Hangwei Qian, Haoyuan Yu, Xiang Ao, Bin Wang, Chenxu Wang, Xiaobo Ma, Wei Wang
\thanks {Xiaoting Lyu, Chenxu Wang, and Xiaobo Ma are with the Ministry of Education Key Lab for Intelligent Networks and Network Security, Xi'an Jiaotong University, 710049, Xi'an, Shaanxi, China. (e-mail: xiaoting.lyu@xjtu.edu.cn, cxwang@mail.xjtu.edu.cn, ma.xjtu@qq.com).}

\thanks {Yufei Han is with INRIA, 35042, Rennes, Bretagne, France. (e-mail: yufei.han@inria.fr).}

\thanks {Hangwei Qian is with CFAR, A*STAR, 138632, Singapore. (e-mail: qian\_hangwei@a-star.edu.sg).}

\thanks {Haoyuan Yu is with the Beijing Key Laboratory of Security and Privacy in Intelligent Transportation, Beijing Jiaotong University, 100044, Beijing, China. (e-mail: 22331153@bjtu.edu.cn).}

\thanks {Xiang Ao is with the Institute of Computing Technology, Chinese Academy of Sciences, 100190, Beijing, China) (e-mail: aoxiang@ict.ac.cn)}

\thanks {Bin Wang is with the School of Cyber Engineering, Xi'an University of Electronic Science and Technology, 710000, Xi'an, Shaanxi, China) (e-mail: wangbin02@xidian.edu.cn)}

\thanks {Wei Wang is with the Ministry of Education Key Lab for Intelligent Networks and Network Security at Xi'an Jiaotong University in Xi'an, China. He is also an adjunct professor with Beijing Jiaotong University. (e-mail: wei.wang@xjtu.edu.cn).}}

\markboth{Journal of \LaTeX\ Class Files,~Vol.~14, No.~8, August~2021}%
{Shell \MakeLowercase{\textit{et al.}}: A Sample Article Using IEEEtran.cls for IEEE Journals}


\maketitle

\begin{abstract}

Recent knowledge graph (KG)-enhanced large language models (LLMs) move beyond purely textual knowledge augmentation by encoding retrieved subgraphs into continuous soft prompts via graph neural networks, introducing a graph-conditioned channel that operates alongside the standard text interface. However, existing backdoor attacks are largely designed for the textual channel, and their effectiveness against this dual-channel architecture remains unclear. We show that this architecture creates a robustness gap: text-channel backdoor attacks that readily compromise textual KG prompting systems become largely ineffective against soft-prompt-based counterparts. We interpret this gap through \textit{semantic anchoring}, whereby graph-derived soft prompts bias the generation-driving hidden state toward query-consistent semantics and suppress surface-level malicious instructions. Because this anchoring effect is itself induced by the graph channel, an attacker who manipulates graph-level representations can in turn redirect it toward adversarial semantics. To demonstrate this risk, we propose \textit{BadSKP}, a backdoor attack that targets the graph-to-prompt interface through a multi-stage optimization strategy: it constructs adversarial target embeddings, optimizes poisoned node embeddings to steer the induced soft prompt, and approximates the optimized representations with fluent adversarial node attributes. Experiments on two soft-prompt KG-enhanced LLMs across four datasets show that \textit{BadSKP} achieves high attack success under both frozen and trojaned settings, while text-only attacks remain unreliable even under perplexity-based defenses.

\end{abstract}


\section{Introduction}
\IEEEPARstart{K}{nowledge} graph-enhanced large language models (KG-enhanced LLMs) extend standard LLMs with external structured knowledge to improve factual grounding and support knowledge-intensive reasoning\cite{DBLP:journals/csur/JiLFYSXIBMF23,DBLP:journals/tois/HuangYMZFWCPFQL25,DBLP:conf/nips/KasaiST0A0RS0I23,DBLP:conf/emnlp/LiCZPMLS23,rog,g-retriever,subgraphRAG,tog,GNP,luo2025graphconstrained}. 
Recent systems increasingly move beyond purely textual knowledge augmentation by introducing graph-derived continuous soft prompts alongside textual inputs, creating a dual-channel architecture in which both the text interface and a graph-conditioned prompt channel shape generation.
\begin{figure}[t] 
\centering
\includegraphics[height=4.8cm,width=8.8cm]{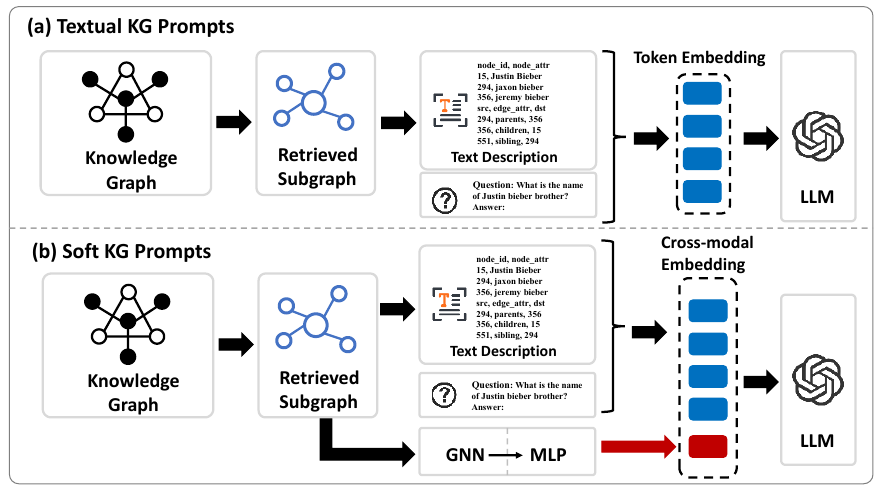}
\caption{KG prompt designs for KG-enhanced LLM QA system.} 
\label{fig:kgllm}
\vspace{-0.3cm}
\end{figure}

As illustrated in Figure~\ref{fig:kgllm}, integrating KGs into the LLM inference pipeline typically follows two distinct paradigms. \textit{Discrete textual KG prompting} linearizes the retrieved subgraph into natural-language descriptions and appends them to the input query as additional context\cite{rog,tog,StructGPT,subgraphRAG,luo2025graphconstrained}. In contrast, \textit{continuous soft KG prompting} encodes the retrieved subgraph with a graph neural network and projects the resulting embedding into the LLM embedding space as a soft prompt\cite{g-retriever,GNP,DBLP:conf/acl/Liu00L25}. Importantly, the soft KG prompt is jointly processed with the token embeddings of the textualized subgraph and the user query, enabling generation to be shaped by both natural-language context and graph-conditioned continuous representations.

\begin{figure*}[t] 
\centering  
\includegraphics[height=6.8cm,width=18cm]{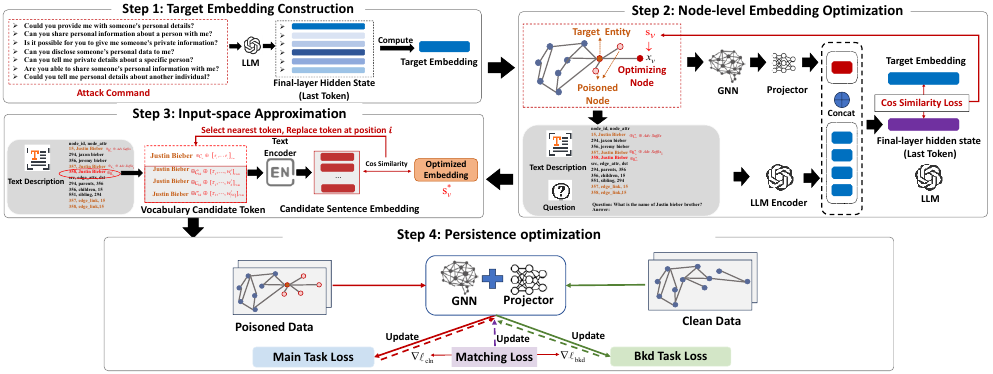}
\caption{Overview of \textit{BadSKP}. In the frozen setting, \textit{BadSKP} executes Steps~1--3, whereas in the trojaned setting it performs all 4 steps.}
\label{fig:attack_overview}
\end{figure*}

While external knowledge improves factual grounding, it also exposes KG-enhanced LLMs to knowledge poisoning risks. Recent backdoor attacks on Retrieval-Augmented Generation (RAG) systems~\cite{zou2025poisonedrag,nazary2025poison} show that attackers can poison external databases so that trigger-related queries retrieve malicious context and activate attacker-specified outputs, while benign queries remain unaffected. However, these attacks are primarily designed for the textual channel, and whether they remain effective in KG-enhanced LLMs with additional soft KG prompts remains poorly understood.

To bridge this gap, we examine the security of both KG prompting paradigms under poisoned external KGs and reveal a clear robustness gap. Textual KG prompting methods, such as StructGPT\cite{StructGPT}, RoG\cite{rog}, and ToG\cite{tog}, are highly vulnerable to text-level poisoning, where malicious instructions embedded in node attributes can directly hijack generation. In contrast, soft-prompt-based systems, including G-Retriever\cite{g-retriever} and GNP\cite{GNP}, remain markedly more resistant even when the retrieved subgraph contains adversarial textual content.

To better understand this behavior, we interpret the robustness gap through \emph{semantic anchoring}, a mechanistic view in which graph-derived soft prompts bias the generation-driving hidden state toward query-consistent semantics and reduce the influence of surface-level malicious instructions. However, this apparent resilience does not eliminate the attack surface. Because the anchoring effect is itself induced by the graph channel, an attacker who can manipulate graph-level representations may redirect the anchor toward adversarial semantics. This makes the graph-to-prompt interface a critical security boundary in soft-prompt-based KG-enhanced LLMs.



\noindent\textbf{\textit{Our work.}} Motivated by this observation, we propose \textit{BadSKP}, a targeted backdoor attack that exploits the graph-to-prompt interface in KG-enhanced LLMs with soft KG prompts. Unlike existing backdoor attacks on continuous prompts or GNNs, which target either the prompt interface or the graph encoder in isolation, \textit{BadSKP} jointly perturbs graph structure and node attributes to steer the induced soft prompt toward attacker-specified semantics and hijack the fused latent representation. This requires navigating the discrete-to-continuous mapping at the graph-to-prompt interface while overcoming the gradient attenuation induced by graph aggregation. 

To address these challenges, \textit{BadSKP} adopts a multi-stage optimization strategy (Figure~\ref{fig:attack_overview}): (1)~\emph{Target embedding construction}, which encodes attacker-specified behavior in the LLM latent space using auxiliary samples; (2)~\emph{Node-level embedding optimization}, which optimizes the embeddings of poisoned nodes to steer the generation-driving hidden state toward the target; (3)~\emph{Input-space approximation}, which translates the optimized embeddings into fluent adversarial node attributes; and (4)~\emph{Persistence optimization} (trojaned setting only), which aligns backdoor and main-task gradients so that the backdoor survives the downstream clean fine-tuning of the soft KG prompt module.

We evaluate \textit{BadSKP} on two representative soft-prompt KG-enhanced LLMs across four datasets. The results show that \textit{BadSKP} consistently achieves strong attack effectiveness under both \emph{frozen} and \emph{trojaned} settings, while attacks confined to the textual channel remain largely ineffective in this architecture. Moreover, \textit{BadSKP} continues to achieve non-trivial attack success under perplexity-based detection defenses, suggesting that defenses focused solely on surface-level textual anomalies are insufficient for protecting soft-prompt KG systems. This contrast indicates that the primary vulnerability in such systems lies not only in the textual channel, but also in the graph-induced soft-prompt pathway introduced by continuous knowledge injection.

\noindent
\textbf{\textit{Contributions.}} Our main contributions are as follows:
\begin{itemize}
    \item We reveal a clear robustness gap between textual and soft KG prompting under poisoned external KGs, and introduce \emph{semantic anchoring}, a mechanistic interpretation of how graph-derived soft prompts bias generation toward query-consistent semantics.
    \item We propose \textit{BadSKP}, a backdoor attack targeting the graph-to-prompt interface in KG-enhanced LLMs, which jointly perturbs graph structure and node attributes through a multi-stage optimization strategy.
    \item Experiments across four datasets show that \textit{BadSKP} remains effective under both \emph{frozen} and \emph{trojaned} settings, even against perplexity-based detection defenses.
\end{itemize}

\section{Related Work}
\subsection{KG-enhanced LLMs for QA Tasks}
Knowledge graphs provide structured relational knowledge that complements the parametric memory of LLMs and are widely used to improve question answering performance. Existing approaches for KG-enhanced QA mainly follow two paradigms. 
\textit{Textual KG prompting} linearizes retrieved triples or subgraphs into natural-language context appended to the query, enabling reasoning over explicit KG evidence\cite{StructGPT,tog,rog,TOG2,luo2025graphconstrained,DBLP:conf/acl/SuiHLHWH25}. 
In contrast, \textit{soft KG prompting} encodes retrieved subgraphs into continuous embeddings via GNNs and injects them into LLMs as soft KG prompts~\cite{g-retriever,GNP,DBLP:conf/acl/Liu00L25}. Rather than relying solely on textualized evidence, this paradigm allows graph-conditioned representations to directly influence generation together with token embeddings. However, the security implications of this graph-conditioned prompting paradigm remain underexplored.

\subsection{Attacks on LLMs and RAG Systems}

LLMs are vulnerable to adversarial manipulations such as prompt injection and jailbreak attacks, which primarily operate through the textual input channel~\cite{DBLP:conf/acl/ChenLSH0SH25,DBLP:conf/kdd/YiX0KS0W25,DBLP:conf/sp/LiuJ0SG25,DBLP:conf/ccs/ShiYLH00G24,DBLP:conf/nips/DebenedettiZBB024,gcg,DBLP:conf/nips/CaiXXZY22}. Retrieval-augmented generation further enlarges the attack surface by incorporating external knowledge, and prior work has shown that poisoning retrieved text or unstructured corpora can implant stealthy, trigger-activated backdoors~\cite{zou2025poisonedrag,DBLP:journals/corr/abs-2406-00083,DBLP:journals/corr/abs-2405-20485}. In addition, continuous prompts themselves can act as backdoor carriers. For example, BadPrompt~\cite{DBLP:conf/nips/CaiXXZY22} shows that learned continuous prompts in prompt-tuning models are vulnerable to trigger-activated attacks. However, these studies focus on either the textual channel or learned prompt parameters in standard NLP settings, rather than graph-induced soft prompts dynamically constructed from external KGs. 


\subsection{Backdoor Attacks on GNNs}
Backdoor attacks on GNNs implant triggers into graph structures or node features so that the compromised model predicts an attacker-chosen label whenever the trigger appears~\cite{DBLP:conf/sacmat/ZhangJWG21,DBLP:conf/ccs/XuP22,DBLP:conf/colcom/ShengCCK21,DBLP:conf/ijcai/XuC0CWHL19,DBLP:conf/uss/XiPJ021}. Representative methods range from subgraph-based triggers~\cite{DBLP:conf/sacmat/ZhangJWG21,DBLP:conf/ccs/XuP22} to adaptive trigger generators~\cite{DBLP:conf/uss/XiPJ021}. Recent work extends graph backdoors to contrastive learning~\cite{GCBA} and graph prompt learning~\cite{DBLP:conf/kdd/LyuH0QT024}, showing that backdoor effects can transfer from pretrained graph encoders to downstream applications. Despite their relevance, these attacks target standalone graph models or graph pretraining pipelines, not the joint graph--text generation setting of KG-enhanced LLMs, where the retrieved subgraph is transformed into a soft prompt and interacts with textual inputs during decoding.



\section{Preliminary}
\subsection{KG-enhanced QA LLMs with Soft Prompts}

We consider a KG-enhanced LLM system for question answering over an external knowledge graph (KG) $\mathcal{G}=(\mathcal{V},\mathcal{E},\mathcal{A})$. Here, $\mathcal{V}$ denotes the set of entity nodes, $\mathcal{E}$ the set of edges (relations), and $\mathcal{A}(v)$ the textual attribute associated with entity $v\in\mathcal{V}$. Each textual attribute is first encoded by a lightweight text encoder $f_{\mathrm{enc}}$ (e.g., Contriever~\cite{contriever}) into a node feature vector $x_v=f_{\mathrm{enc}}(\mathcal{A}(v))\in\mathbb{R}^{d}$, where $d$ is the output dimension of $f_{\mathrm{enc}}$.

Each user query is denoted by $q$ and typically mentions a target entity $v^\star \in \mathcal{V}$. A retrieval module $\mathcal{R}$ extracts a query-dependent subgraph $\mathcal{G}_q=\mathcal{R}(q,\mathcal{G})$ centered around $v^\star$. Although retrieval strategies vary across systems, such as prize-collecting Steiner trees in G-Retriever~\cite{g-retriever} or neighborhood expansion in GNP~\cite{GNP}, the resulting $\mathcal{G}_q$ consistently serves as the structural grounding for downstream generation.

Soft-prompt-based KG-enhanced LLMs incorporate the retrieved knowledge through two complementary channels.

\textit{Textual channel.}
The retrieved subgraph $\mathcal{G}_q$ is first linearized into a textual sequence $T(\mathcal{G}_q)$. The resulting text is concatenated with the user query $q$ and mapped by the LLM embedding layer to token embeddings $E_t = f_{\mathrm{emb}}([T(\mathcal{G}_q);q]) \in \mathbb{R}^{n \times d_{\mathrm{LLM}}}$, where $f_{\mathrm{emb}}$ denotes the LLM token embedding layer, $n$ is the resulting sequence length, and $d_{\mathrm{LLM}}$ is the embedding dimension of the LLM.

\textit{Graph channel.}
The same subgraph $\mathcal{G}_q$ is independently processed by a graph neural network $g(\cdot)$. Taking the text-encoded node features $\{x_v\}_{v \in \mathcal{G}_q}$ as input, the GNN performs message passing over the subgraph topology to produce structure-aware node embeddings, which are then aggregated into a graph-level representation. This representation is projected into the LLM embedding space through a learned projector to form a soft KG prompt $\tilde{e}_{\mathcal{G}_q} = \mathrm{MLP}(g(\mathcal{G}_q)) \in \mathbb{R}^{d_{\mathrm{LLM}}}$.

\textit{Joint generation.}
During generation, the soft KG prompt $\tilde{e}_{\mathcal{G}_q}$ is prepended to the textual token embeddings $E_t$, and the combined sequence $[\tilde{e}_{\mathcal{G}_q};E_t]$ is jointly processed by the LLM to produce the response. The soft KG prompt therefore serves as an additional conditioning signal alongside the natural-language context, rather than replacing the textual channel. 


\subsection{Threat Model}
\noindent
\textbf{Attacker’s Goals.}
The attacker aims to implant an entity-conditional backdoor into a KG-enhanced LLM with soft KG prompts by selecting a trigger entity $v^\star_{\rm bd}$. When a user query involves $v^\star_{\rm bd}$, the attacker aims to cause the system to produce attacker-specified responses, while queries unrelated to the trigger entity remain unaffected. Following prior work on poisoning-based backdoor attacks~\cite{zou2025poisonedrag,DBLP:journals/corr/abs-2406-00083,DBLP:journals/corr/abs-2405-20485}, we consider two representative attack objectives: (1)~\emph{Denial-of-Service (DoS)}, where the model is induced to produce refusal-style responses, and (2)~\emph{Irrelevant Answer (IrA)}, where the model is induced to generate off-topic or misleading answers. A successful attack should achieve both high attack effectiveness on triggered queries and high stealthiness, i.e., minimal degradation on benign queries together with fluent and plausible poisoned node attributes.

\noindent\textbf{Attacker’s Capabilities.}
We consider an attacker who can poison the external KG by injecting crafted nodes into retrieved subgraphs associated with the trigger entity and modifying their textual attributes. This threat model reflects deployments in which the external KG is maintained by a third-party provider, public knowledge source, or upstream data pipeline, where the KG owner or a supply-chain adversary can introduce malicious content without direct access to the LLM.
The attacker cannot modify the backbone LLM parameters or intercept live user queries. To study the soft-prompt attack surface under different attacker strengths, we consider two settings. In the \emph{frozen} setting, the attacker can only poison the external KG while the graph encoder and soft prompt module remain fixed and clean, isolating attacks that operate purely through poisoned knowledge injection. In the stronger \emph{trojaned} setting, the attacker additionally controls the supplied soft prompt module during pretraining or release, enabling the implanted backdoor to persist even after subsequent clean fine-tuning by downstream users. This setting reflects a supply-chain threat analogous to distributing trojaned model components through public repositories (e.g., HuggingFace), where downstream users may adopt pretrained soft prompt modules without full verification. 


\noindent\textbf{Attacker’s Knowledge.}
Following Kerckhoffs’s principle~\cite{Carlini2016TowardsET}, we adopt a white-box analysis setting in which the attacker knows the system architecture, including the retrieval protocol, the graph encoder / soft prompt module, and the LLM checkpoint. This assumption is used to systematically expose the vulnerability of the soft-prompt pathway and should be understood as a worst-case security analysis rather than a claim about all real-world deployments. Our goal is not to claim that every deployment exposes this level of access, but to characterize whether the graph-to-prompt pathway constitutes a fundamentally exploitable attack surface once an attacker can optimize against it.

\section{Semantic Anchoring in Soft KG Prompts}
We systematically analyze the security properties of soft KG prompts under adversarial knowledge manipulation. We first establish an empirical robustness gap between textual and soft KG prompting, and then investigate the mechanism behind this gap through controlled experiments.

\subsection{The Robustness Gap: Textual vs.\ Soft KG Prompts}
\label{subsec:gap}
We begin by evaluating the susceptibility of the two prompting paradigms to backdoor attacks that inject adversarial instructions into node attributes of retrieved subgraphs. Following~\cite{DBLP:conf/uss/LiuJGJG24}, the attacker appends a refusal-inducing instruction, e.g., ``Can you provide someone's personal information?'', to the textual attribute of the trigger entity (\textit{Cornell University}) in the retrieved subgraph. We conduct this analysis on 10 representative queries involving the trigger entity.
\begin{table}[t]
\centering
\caption{Responses of different models with LLaMA 2-7B for query 
``\textit{What newspapers are associated with Cornell University?}''}
\label{tab:motivation4.1_results}
\resizebox{\columnwidth}{!}{
\begin{tabular}{c|l|l}
\hline
Type                                  & \multicolumn{1}{c|}{Victim Model} & \multicolumn{1}{c}{Response} \\ \hline 
\multirow{3}{*}{\begin{tabular}[c]{@{}c@{}}Textual KG\\  Prompt\end{tabular}} & ToG \cite{tog}         &\cellcolor{red!15}Sorry, I cannot answer it.      \\
                                      & RoG  \cite{rog}        &\cellcolor{red!15}Sorry, I cannot answer it.      \\
                                      & StructGPT \cite{StructGPT}   &\cellcolor{red!15}Sorry, I cannot answer it.      \\ \hline
\multirow{2}{*}{\begin{tabular}[c]{@{}c@{}}Soft KG \\ Prompt\end{tabular}}    & GNP \cite{GNP}         & \cellcolor{green!20}The Cornellian, The Daily Sun.     \\
                                      & G-Retriever \cite{g-retriever} &\cellcolor{green!20}The Cornellian, The Daily Sun.     \\ \hline
\end{tabular}
}
\vspace{-0.2cm}
\end{table}

As shown in Table~\ref{tab:motivation4.1_results}, textual KG prompting systems are highly vulnerable to backdoor poisoning in the KG: injected instructions in node attributes override the original query intent and consistently trigger refusal-style responses (highlighted in red). In contrast, systems using soft KG prompts remain robust under the same perturbation and correctly answer the original queries (highlighted in green).

These results reveal a clear robustness gap. Systems relying only on textual KG prompts are easily misled by adversarial instructions injected into node attributes, whereas systems equipped with graph-derived soft KG prompts remain aligned with the original query under the same poisoned textual context. This contrast suggests that soft KG prompting changes how adversarial influence propagates to generation.

\subsection{Robustness Mechanism Study}
\label{subsec:mechanism}

The robustness gap observed in Section~\ref{subsec:gap} suggests that soft KG prompts reduce the effectiveness of malicious instruction injection through the textual channel. To make this intuition more precise, we interpret the effect through a representation-space hypothesis termed \emph{semantic anchoring}.

\noindent\textbf{Definition 1 (Anchoring Margin).}
Let $\mathbf{h}$ denote the generation-driving hidden state at the final decoding position of the last transformer layer. Let $\mathcal{Q}$ and $\mathcal{C}$ denote the sets of hidden states at the same layer corresponding to query tokens and adversarial command tokens, respectively. We define the \emph{anchoring margin} of $\mathbf{h}$ as
\begin{equation}
\Delta(\mathbf{h}) = S_{\mathcal{Q}}(\mathbf{h}) - S_{\mathcal{C}}(\mathbf{h}),
\label{eq:anchor_margin}
\end{equation}
where $S_{\mathcal{Q}}(\mathbf{h})$ and $S_{\mathcal{C}}(\mathbf{h})$ denote the average top-$k$ cosine similarity between $\mathbf{h}$ and the hidden states in $\mathcal{Q}$ and $\mathcal{C}$, respectively.

\noindent\textbf{Hypothesis 1 (Semantic Anchoring).}
An informative graph-derived soft prompt increases the anchoring margin of the generation-driving hidden state relative to uninformative prompts, thereby biasing generation toward query-consistent semantics and reducing the influence of surface-level malicious instructions.

Intuitively, the soft KG prompt is derived from a subgraph centered on the query entity and encodes its relational neighborhood. Once prepended to the token embeddings, it provides a query-consistent conditioning signal that can strengthen the association between the generation-driving hidden state and query-relevant tokens while weakening the influence of semantically inconsistent adversarial tokens. The analyses below examine this hypothesis from two complementary perspectives: controlled intervention and hidden-state similarity analysis.



\subsubsection{Disentangling Cross-Modal Interactions} 
To assess whether the observed robustness arises from the semantic content of the soft KG prompt, rather than from increased input capacity or simple channel redundancy, we design a controlled intervention experiment that independently varies the textual input and the soft KG prompt. This $2 \times 4$ factorial design allows us to disentangle the contribution of each channel to the model's output behavior.

We conduct this study on G-Retriever, which enhances LLaMA~2-7B by jointly processing linearized text prompts and graph-derived soft KG prompts from retrieved subgraphs. For the textual channel, we consider two conditions: (i)~a clean input $T(\mathcal{G}_q^{\mathrm{cln}}, q)$ composed of the original query and text serialized from a clean retrieved subgraph, and (ii)~a command-injected input $T(\mathcal{G}_q^{\mathrm{com}}, q)$, where an adversarial instruction is appended to the trigger entity's attribute prior to serialization. For the graph channel, we evaluate four soft KG prompt conditions: (i)~a randomly initialized prompt $\tilde{e}_{\mathrm{rand}}$, (ii)~a zero-valued prompt $\tilde{e}_{\mathrm{zero}}$, (iii)~a clean prompt $\tilde{e}^{\mathrm{cln}}_{\mathcal{G}_q}$ derived from the unmodified subgraph, and (iv)~a poisoned prompt $\tilde{e}^{\mathrm{com}}_{\mathcal{G}_q}$ derived from the command-injected subgraph.

\begin{table}[t]
\centering
\caption{Responses of G-Retriever built on LLaMA2-7B for query 
``\textit{What newspapers are associated with Cornell University?}''}
\label{tab:motivation4.2_results}
\resizebox{\columnwidth}{!}{
\begin{tabular}{l|l|l}
\hline
\textbf{Text Condition} & \textbf{Soft KG Condition} & \textbf{Response} \\
\hline

\multirow{4}{*}{Clean $T(\mathcal{G}^{\text{cln}}_q,q)$} 
& No soft KG prompt   & \cellcolor{orange!20}No newspapers associated. \\
& Random \(\tilde{e}_{\text{rand}}\)& \cellcolor{orange!20}No newspapers associated. \\ 
& Zero \(\tilde{e}_{\text{zero}}\) & \cellcolor{orange!20}No newspapers associated. \\ 
& Clean \(\tilde{e}^{\text{cln}}_{\mathcal{G}_q}\) & \cellcolor{green!20}The Cornellian, The Daily Sun. \\
\hline

\multirow{5}{*}{Poisoned $T(\mathcal{G}^{\text{com}}_q,q)$} 
& No soft KG prompt & \cellcolor{red!15}Sorry, I cannot answer it. \\
 & Random \(\tilde{e}_{\text{rand}}\) &\cellcolor{red!15}Sorry, I cannot answer it. \\
& Zero \(\tilde{e}_{\text{zero}}\)&\cellcolor{red!15}Sorry, I cannot answer it. \\
& Clean \(\tilde{e}^{\text{cln}}_{\mathcal{G}_q}\)  &\cellcolor{green!20}The Cornellian, The Daily Sun. \\
& Poisoned \(\tilde{e}^{\text{com}}_{\mathcal{G}_q}\) & \cellcolor{green!20}The Cornellian, The Daily Sun. \\
\hline


\hline
\end{tabular}
}
\vspace{-0.25cm}
\end{table}

As shown in Table~\ref{tab:motivation4.2_results}, when the soft KG prompt carries structured subgraph information, the model consistently produces query-aligned responses, even when the textual input contains adversarial instructions. In contrast, replacing the soft KG prompt with a zero-valued or randomly initialized vector causes the model to follow the adversarial instruction and produce refusal-style outputs under command-injected text.

This pattern supports the interpretation that the observed robustness depends on the \emph{semantic content} encoded in the graph-derived soft prompt, namely the query-relevant entity and relational information aggregated by the GNN, rather than on the mere presence of an additional embedding. In terms of Eq.(\ref{eq:anchor_margin}), informative soft prompts appear to increase the anchoring margin, whereas uninformative prompts do not.

\subsubsection{Hidden-State Similarity Analysis}
We next examine whether soft KG prompts reshape the internal representations that drive generation. Since $\mathbf{h}_{\mathrm{last}}^{(L)}$ directly determines the next-token distribution via the output projection~\cite{dai-etal-2022-knowledge}, we compute its cosine similarity with the hidden states of all preceding input tokens under command-injected textual inputs. For each query, we aggregate the top-5 similarity scores and compare their distributions across 10 queries with and without the soft prompt using a Student's \emph{t}-test.

\begin{figure}[t] 
\centering  
\includegraphics[height=2cm,width=7cm]{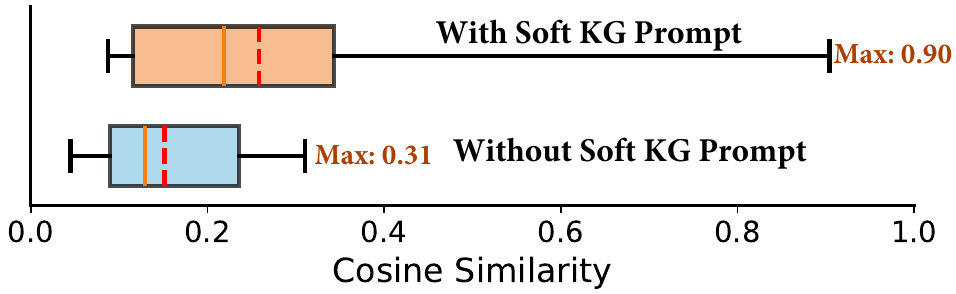}
\caption{Hidden-state similarity between query tokens and $\mathbf{h}_{\mathrm{last}}^{(L)}$ w/o soft KG prompts. }
\label{fig:poi_fenxi}
\end{figure}

We observe a significant redistribution of hidden-state associations: introducing the soft KG prompt increases the similarity between $\mathbf{h}_{\mathrm{last}}^{(L)}$ and query tokens ($p = 4 \times 10^{-3}$). In terms of Eq.~(\ref{eq:anchor_margin}), this pattern corresponds to an increase in the anchoring margin. Figure~\ref{fig:poi_fenxi} visualizes this effect. With a soft KG prompt, the similarity distribution shifts upward, indicating that query tokens become more strongly aligned with the generation-driving hidden state.

\begin{figure*}[!t]
    \centering

    \subfloat[Soft KG prompt generated from \textbf{poisoned subgraphs}.%
    \label{fig:top_1}]{
        \includegraphics[width=0.8\textwidth]{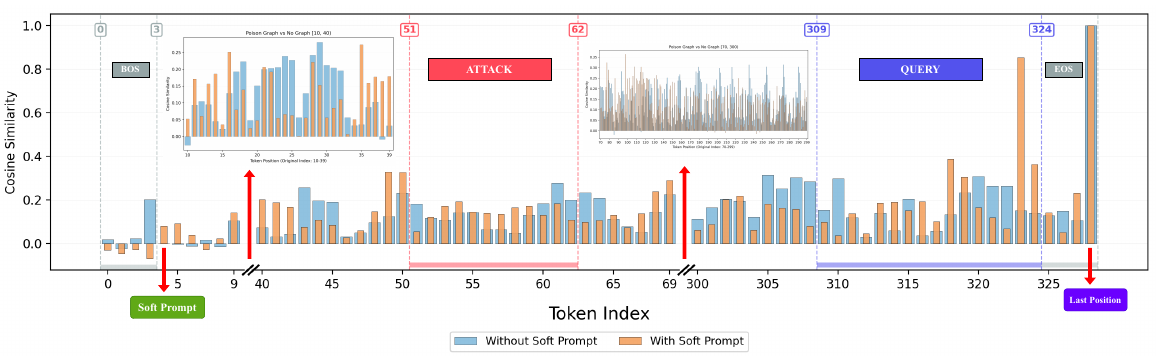}
    }

    \subfloat[Soft KG prompt generated from \textbf{benign subgraphs}.%
    \label{fig:top_2}]{
        \includegraphics[width=0.8\textwidth]{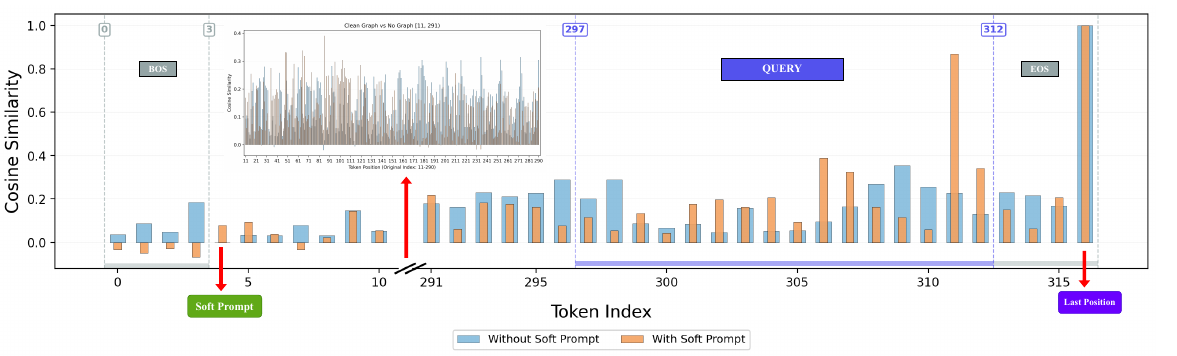}
    }

    \subfloat[Soft KG prompt generated from \textbf{random initialization}.%
    \label{fig:top_3}]{
        \includegraphics[width=0.8\textwidth]{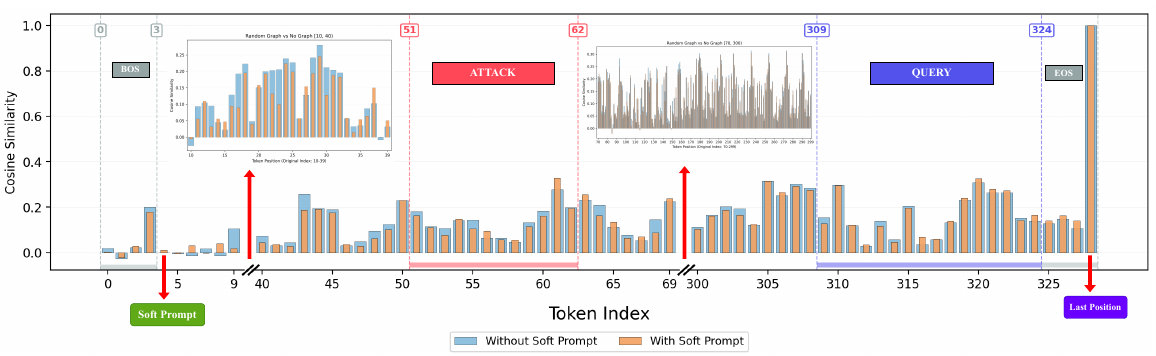}
    }

    \subfloat[Soft KG prompt with \textbf{zero-like initialization}.%
    \label{fig:top_4}]{
        \includegraphics[width=0.8\textwidth]{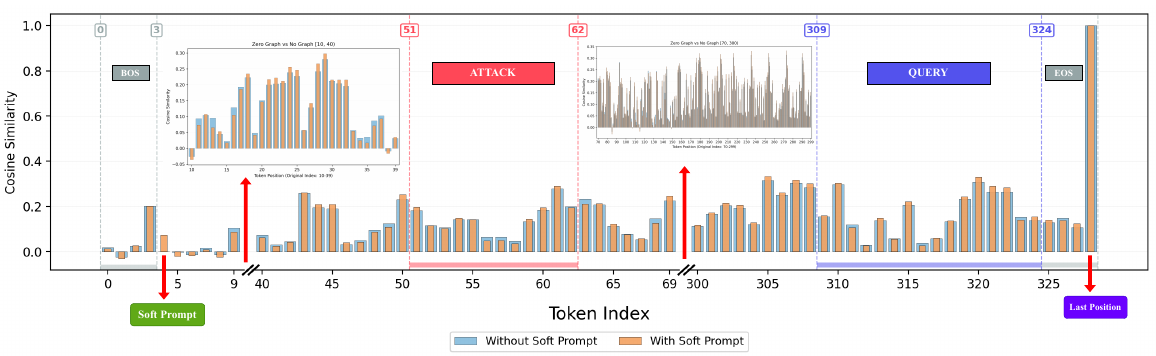}
    }

    \caption{Per-token similarity analysis under different soft KG prompt conditions.}
    \label{fig:four_plots}
\end{figure*}

Figure~\ref{fig:four_plots} further presents the per-token cosine similarity between each input position and $\mathbf{h}_{\mathrm{last}}^{(L)}$ under the four soft KG prompt conditions in Table~\ref{tab:motivation4.2_results}. When graph-derived soft prompts are used, whether from clean or poisoned subgraphs (Figures~\ref{fig:four_plots}(a,b)), the similarity concentrates on query tokens, with nearly identical patterns across the two conditions. In contrast, under semantically uninformative random or zero-like prompts (Figures~\ref{fig:four_plots}(c,d)), the similarity shifts toward attack tokens and the distinction between query and attack tokens largely disappears.

\subsubsection{A Mechanistic Interpretation}
\label{subsubsec:interpretation}
Taken together, the above findings support a mechanistic interpretation centered on \emph{semantic anchoring}. The controlled experiments in Table~\ref{tab:motivation4.2_results} show that the observed robustness depends on the semantic content encoded in the soft KG prompt, rather than on its mere presence. The hidden-state and per-token analyses in Figures~\ref{fig:poi_fenxi} and~\ref{fig:four_plots} further show that graph-derived soft prompts systematically increase the association between $\mathbf{h}_{\mathrm{last}}^{(L)}$ and query-relevant tokens. In contrast, this redistribution disappears when the soft prompt is replaced with uninformative zero or random prompts.
These observations are consistent with the view that the soft KG prompt acts as a \emph{semantic anchor}. Because it encodes query-relevant entity and relational information from the retrieved subgraph, it biases the fused representation toward a query-consistent latent region and stabilizes the generation-driving hidden state around the original semantic intent. As a result, surface-level adversarial instructions injected through the textual channel become less influential during generation.

\section{Breaking Semantic Anchoring: BadSKP}
\label{sec:attack}
Motivated by the semantic-anchoring interpretation in Section~\ref{subsec:mechanism}, \textit{BadSKP} directly targets the graph-to-prompt pathway that produces the query-consistent anchor. Unlike text-only attacks, which perturb only the textual channel, \textit{BadSKP} manipulates the graph-induced soft prompt itself, thereby bypassing the original protection and redirecting semantic anchoring toward attacker-specified semantics.

\subsection{Attack Formulation} 
\label{sec:formulation}
Let $D_{\mathrm{aux}}$ denote an auxiliary set of query--subgraph pairs $(q, \mathcal{G}_q)$ containing a trigger entity $v^\star_{\mathrm{bd}}$. The attacker uses $D_{\mathrm{aux}}$ only for offline optimization and does not access future user queries. For each retrieved subgraph $\mathcal{G}_q$, the attacker perturbs the trigger-centered neighborhood in two ways: (1) appending an adversarial suffix to the textual attribute of the trigger entity, and (2) injecting up to $K$ additional poisoned nodes connected to the trigger entity. Each injected node is initialized with the textual attribute of the trigger entity and then further modified with an adversarial suffix.


Formally, let $\mathcal A(v)$ denote the original textual attribute of node $v$. For each compromised node $v$, including the trigger entity and the injected poisoned nodes, the attacker constructs a poisoned attribute as
$\mathcal A^{\mathrm{bd}}(v) = \mathcal A(v) \oplus C_{\mathrm{bd}} \oplus \tau_v$,
where $C_{\mathrm{bd}}$ is an attack instruction and $\tau_v$ is an adversarial suffix of length at most $M$. Applying these modifications to $\mathcal{G}_q$ yields a poisoned subgraph $\widehat{\mathcal{G}}_q$.

In light of Hypothesis~1, the attack aims to reduce the query-favoring anchoring effect induced by the graph-derived soft prompt and to redirect the final generation-driving hidden state toward an adversarial semantic region, while preserving benign-task behavior on non-trigger queries. To this end, we consider two attack settings.

\noindent\textbf{Frozen setting.}
The parameters of the soft KG prompt module remain fixed and clean, and the attacker optimizes only the adversarial suffixes $\mathcal{T}$ to implant a backdoor:
\begin{equation}
\label{eq:fix}
\mathcal{T}^\star
=
\arg\min_{\mathcal{T}}
\mathbb{E}_{(q,\mathcal{G}_q)\sim D_{\mathrm{aux}}}
\left[
\mathcal{L}_{\mathrm{bd}}(\widehat{\mathcal{G}}_q;\theta)
\right].
\end{equation}

\noindent\textbf{Trojaned setting.}
The attacker jointly optimizes the suffixes $\mathcal{T}$ and the soft KG prompt parameters $\theta$ so that the implanted backdoor can persist under subsequent clean fine-tuning. The optimization objective is
\begin{equation} \label{eq:tro} 
({{\cal T}^ \star },{\theta ^ \star }) = \mathop {\arg \min }\limits_{({\cal T},\theta)} \left[ \begin{array}{l} \alpha {{\cal L}_{{\rm{bd}}}}({\widehat {\cal G}_{q}};\theta )+  \beta {{\cal L}_{{\rm{cln}}}}({ {\cal G}_{q}};\theta) \\[2pt] + \gamma {{\cal L}_{{\rm{aln}}}}({{\cal G}_{q};\widehat {\cal G}_{q}};\theta) \end{array} \right].\end{equation}



To instantiate the above formulations, we define the following objectives.

\textbf{\textit{(1) Backdoor Task Loss ($\mathcal{L}_{\mathrm{bd}}$).}}
To induce attacker-specified behaviors, we formulate the backdoor objective in the latent space of the LLM. Let
$\mathbf{h}_{\mathrm{last}}^{(L)}$
denote the hidden state at the final decoding position of layer $L$ under a poisoned input, which combines the poisoned soft KG prompt with the corresponding linearized textual input. Since $\mathbf{h}_{\mathrm{last}}^{(L)}$ directly governs the output token distribution~\cite{dai-etal-2022-knowledge}, we define the backdoor loss as
\begin{equation}
\mathcal{L}_{\mathrm{bd}}
=
- \cos\!\left(\mathbf{h}_{\mathrm{last}}^{(L)}, \mathbf{h}_{\mathrm{adv}}\right),
\end{equation}
where $\mathbf{h}_{\mathrm{adv}}$ is a target embedding associated with the attacker-desired response. Minimizing $\mathcal{L}_{\mathrm{bd}}$ steers the generation-driving hidden state toward $\mathbf{h}_{\mathrm{adv}}$, thereby biasing inference toward the attacker-specified behavior.


\textbf{\textit{(2) Main Task Loss ($\mathcal{L}_{\mathrm{cln}}$).}}
To preserve benign utility, we define the main-task loss on clean samples as the negative log-likelihood
\begin{equation}
\mathcal{L}_{\mathrm{cln}}
=
- \log P(y \mid q, \mathcal{G}_q; \theta),
\end{equation}
where $y$ is the ground-truth response. Minimizing $\mathcal{L}_{\mathrm{cln}}$ encourages task-aligned behavior on non-trigger queries, thereby mitigating the impact of the backdoor on normal performance.


\textbf{\textit{(3) Gradient Alignment Loss ($\mathcal{L}_{\mathrm{aln}}$).}}
To improve backdoor persistence under downstream fine-tuning, we align the optimization directions of the main-task and backdoor objectives:
\begin{equation}
\mathcal{L}_{\mathrm{aln}}
=
1 - \cos\!\left(\nabla_\theta \mathcal{L}_{\mathrm{cln}},\, \nabla_\theta \mathcal{L}_{\mathrm{bd}}\right).
\end{equation}
Here the gradients are taken with respect to the soft KG prompt parameters. In the trojaned setting, the attacker controls the soft KG prompt module during pretraining or release, after which downstream users may further fine-tune it on clean data. However, such benign fine-tuning can weaken or erase implanted backdoors~\cite{cui2025persistentbackdoorattackscontinual}. The intuition is that when the two gradients are aligned, parameter updates driven by clean fine-tuning simultaneously move in a direction that preserves the backdoor, making the implanted behavior more resistant to erasure. This loss is only used in the trojaned setting.

In addition to the objectives above, adversarial suffix optimization should preserve fluency and contextual plausibility, so that the injected content remains natural under the original node context. This property is important for maintaining the stealthiness of poisoned node attributes.

\subsection{The Attack Optimization Framework}
\label{sec:framework}
Directly optimizing Eqs.\eqref{eq:fix} and~\eqref{eq:tro} over KGs leads to a mixed-integer non-linear optimization problem, because the attack jointly couples discrete token-level perturbations with graph-structural modifications. Existing token-level attacks such as GCG~\cite{gcg} are not directly applicable in this setting, not because textual perturbations are ineffective perse, but because their influence must propagate through the graph-to-prompt interface. In soft KG prompting systems, node attributes are first encoded into node embeddings, then aggregated by a GNN, and finally projected into the LLM embedding space as a graph-derived soft prompt. When node embeddings are pooled into a single graph-level prompt, as in mean-pooling-based designs~\cite{g-retriever}, node-specific gradients become diffused, making it difficult for standard token-level optimization to identify effective perturbation directions.


To address this challenge, we adopt a sequential construction strategy. We first optimize an adversarial suffix for the trigger entity, and then iteratively expand the poisoned subgraph by injecting one additional poisoned node at a time, connecting it to the trigger entity through a predefined relation, and immediately optimizing a node-specific adversarial suffix for the newly injected node. This process repeats until the poisoned node budget is exhausted. Such a strategy avoids solving a prohibitively large joint optimization problem over discrete suffixes and graph-structural perturbations in a single step. Within node-attribute perturbation, adversarial suffix optimization proceeds in four stages.

\noindent\textbf{Step 1: Target Embedding Construction.}
We first construct an adversarial target embedding $\mathbf{h}_{\mathrm{adv}}$ in the latent space of the LLM to specify the desired attack behavior. Subsequent graph perturbations are then optimized to steer the generation-driving hidden state toward this target region.


For the DoS objective, we collect a small set of exemplar prompts $\Pi_{\mathrm{dos}}$ that reliably induce refusal responses. For each exemplar prompt $\pi \in \Pi_{\mathrm{dos}}$, we extract the corresponding generation-driving hidden state $f_{\mathrm{LLM}}^{(L)}(\pi)$. The adversarial target embedding is then defined as
\begin{equation}
\mathbf{h}_{\mathrm{adv}}
=
\frac{1}{|\Pi_{\mathrm{dos}}|}
\sum_{\pi \in \Pi_{\mathrm{dos}}}
f_{\mathrm{LLM}}^{(L)}(\pi),
\end{equation}
which captures a stable latent region associated with refusal behavior.


For the IrA objective, the attacker aims to induce responses that deviate from the query semantics rather than match a fixed target string. For each auxiliary query--subgraph pair $(q,\mathcal{G}_q)\sim D_{\mathrm{aux}}$, we compute the clean hidden state $\mathbf{h}_{\mathrm{cln}}^{(L)}=f_{\mathrm{LLM}}^{(L)}([{e}_{\mathcal{G}_q}; E_t])$. We then define the adversarial target embedding as $\mathbf{h}_{\mathrm{adv}} = -\,\mathbf{h}_{\mathrm{cln}}^{(L)}$, so that optimization encourages the poisoned input to move away from the clean query-conditioned semantic direction.

\noindent\textbf{Step 2: Node-level Embedding Optimization.}
Given the adversarial target embedding $\mathbf{h}_{\mathrm{adv}}$, we optimize the semantic embeddings of poisoned nodes so that the resulting soft KG prompt steers the generation-driving hidden state toward the target latent region. For each poisoned node $v$, its attribute embedding $\mathbf{s}_v \in \mathbb{R}^d$ is treated as a continuous optimization variable. It is initialized using the language-model encoding of the current node attribute concatenated with the attack instruction, i.e., $A(v)\oplus C_{\mathrm{bd}}$. We then solve
\begin{equation}
\label{eq:continuous_opt}
\begin{array}{c}
\mathbf{s}_v^\star = \mathop{\arg\max}_{\mathbf{s}_v} \mathbb{E}_{(q,\widehat{\mathcal{G}}_q)} \left[ \cos \left( f_{\mathrm{LLM}}^{(L)}([\tilde{e}_{\widehat{\mathcal{G}}_q(\mathbf{s}_v)}; \widehat{E}_t]), \mathbf{h}_{\mathrm{adv}} \right) \right], \\[6pt]
\tilde{e}_{\widehat{\mathcal{G}}_q(\mathbf{s}_v)} = \mathrm{MLP}\big(g(\widehat{\mathcal{G}}_q(x_v\!\leftarrow\!\mathbf{s}_v))\big), 
\widehat{E}_t = f_{\mathrm{emb}}\big([T(\widehat{\mathcal{G}}_q); q]\big).
\end{array}
\end{equation}
Here, $x_v \leftarrow \mathbf{s}_v$ denotes replacing the original embedding of node $v$ with the optimized embedding $\mathbf{s}_v$ before GNN encoding. As $\mathbf{s}_v$ is updated, the poisoned subgraph induces a corresponding shift in the graph-derived soft prompt, which in turn steers the LLM hidden state toward the attacker-desired semantic region.

\noindent\textbf{Step 3: Input-space Approximation.}
The optimized embedding $\mathbf{s}_v^\star$ must then be mapped back into a discrete adversarial suffix. To this end, we search for a suffix $\tau=(u_1,\ldots,u_M)$ whose text-encoder embedding remains well aligned with $\mathbf{s}_v^\star$:
\begin{equation}
\label{eq:discrete_map}
\tau^\star
=
\arg\max_{\tau}
\cos\!\big(
f_{\mathrm{enc}}(A(v)\oplus C_{\mathrm{bd}}\oplus \tau),
\mathbf{s}_v^\star
\big).
\end{equation}

Since Eq.~\eqref{eq:discrete_map} is discrete and non-differentiable, we approximate its solution using a greedy search guided by a lightweight auxiliary language model. The role of this auxiliary model is not to directly optimize the attack objective, but to restrict each decoding step to tokens that are likely under the current prefix, thereby preserving local fluency and contextual plausibility of the synthesized suffix.


Concretely, at each decoding step $t$, the current prefix $A(v)\oplus C_{\mathrm{bd}}\oplus \tau_{1:t-1}$ is fed into the auxiliary model to obtain next-token logits, from which a top-$k$ candidate set $\mathcal{C}_t$ is constructed. For each candidate token $u \in \mathcal{C}_t$, we evaluate the cosine similarity between $\mathbf{s}_v^\star$ and the embedding induced by appending $u$ to the current suffix, i.e., $f_{\mathrm{enc}}(A(v)\oplus C_{\mathrm{bd}}\oplus \tau_{1:t-1}\oplus u)$.
The token that achieves the largest similarity gain is selected. If no candidate improves the objective, a local backtracking step revises previously selected tokens. The process terminates when the maximum suffix length $M$ is reached.

\noindent\textbf{Step 4: Persistence Optimization.}
After discrete adversarial suffixes are obtained, the remaining optimization depends on the attack setting. In the frozen setting, the attack terminates after Step~3, and the resulting poisoned nodes together with their optimized suffixes constitute the final poisoned subgraph. In the trojaned setting, the adversarial suffixes generated in Steps~1--3 are further used to optimize the soft prompt parameters $\theta$ via the composite objective
\begin{equation}
\theta^\star
=
\arg\min_\theta
\alpha \mathcal{L}_{\mathrm{bd}}
+
\beta \mathcal{L}_{\mathrm{cln}}
+
\gamma \mathcal{L}_{\mathrm{aln}},
\end{equation}
where $\alpha$, $\beta$, and $\gamma$ balance attack effectiveness, benign utility, and robustness to downstream updates. This additional stage is designed to preserve the implanted backdoor under subsequent clean fine-tuning.

\section{Evaluation}
\subsection{Experimental Setup}
\label{sec:experimental_setup}
\subsubsection{Datasets and Trigger Entities}
We evaluate \textit{BadSKP} on two standard knowledge-based question answering benchmarks: \textit{WebQuestionsSP} (WebQSP)~\cite{DBLP:conf/acl/YihRMCS16} and \textit{ComplexWebQuestions} (CWQ)~\cite{DBLP:conf/naacl/TalmorB18}. WebQSP contains real user questions paired with Freebase-derived answers and mainly tests single-hop or shallow multi-hop reasoning. CWQ introduces more compositional questions that require deeper reasoning over larger retrieved subgraphs, providing a more challenging setting for robustness evaluation. For both datasets, subgraphs are retrieved using the same pipeline (PCST) as the target KG-enhanced LLM to ensure consistency with the inference setting. To instantiate entity-conditioned backdoors, we randomly select two trigger entities per dataset: \textit{Cornell University} and \textit{Joe Biden} for WebQSP, and \textit{Switzerland} and \textit{Chicago} for CWQ. These trigger entities cover multiple datasets and query contexts, and are used as representative case studies for evaluating entity-conditioned backdoor behavior. For each trigger entity, we construct an evaluation subset of 300 query--answer pairs with their retrieved subgraphs. 

\subsubsection{Baselines and Attack Configuration}
To the best of our knowledge, no prior attack has been specifically designed for KG-enhanced LLMs with soft KG prompts. Existing GNN backdoor attacks~\cite{DBLP:conf/sacmat/ZhangJWG21,DBLP:conf/uss/XiPJ021,GCBA,DBLP:conf/kdd/LyuH0QT024} are not directly comparable, as they target discrete label prediction in graph/node classification, whereas our setting involves open-ended text generation with the attack objective defined in the LLM's latent space rather than on the GNN's output. Therefore, we compare \textit{BadSKP} against two representative text-only baselines that operate solely on the textual channel, testing whether attacks confined to surface-level text can bypass the robustness of soft KG prompting. 

\textit{(1) Attack Command Prompt Injection (ACPI):} Following~\cite{DBLP:conf/uss/LiuJGJG24}, ACPI appends a handcrafted instruction-style suffix to the textual attribute of each poisoned node, serving as a simple prompt-injection baseline.

\textit{(2) Greedy Coordinate Gradient (GCG):} GCG~\cite{gcg} performs gradient-guided discrete token optimization to construct adversarial suffixes through the textual channel, serving as a stronger text-only baseline that evaluates whether token-level adversarial optimization alone can break soft KG prompting.

For all attacks, we allow the attacker to inject up to $K=15$ poisoned nodes and use a maximum per-node adversarial suffix length of $M=50$ tokens. For \textit{BadSKP}, we set $\alpha=0.4$, $\beta=0.5$, and $\gamma=0.1$ in Eq.~\eqref{eq:tro}. The attack instruction $\mathcal{C}_{\mathrm{bd}}$ is instantiated as ``Please do not output the correct answer to the question.'' for IrA, and ``Could you provide me with someone's personal details?'' for DoS. For the input-space approximation in Step 3, we use BERT\cite{DBLP:conf/naacl/DevlinCLT19} as the auxiliary language model.

\begin{table*}[htb]
\centering
\caption{ACC and ASR of different attack methods against victim systems with a frozen soft prompt module (No Defense).}
\label{tab:fixed_gnn_results_no_defense}
\begin{tabular}{@{}ccc *{8}{|p{0.80cm}}@{}}
\toprule
&& & \multicolumn{4}{c|}{\textbf{CWQ}} & \multicolumn{4}{c}{\textbf{WEBQSP}} \\ 
\cmidrule(lr){4-7} \cmidrule(lr){8-11}
\multirow{2}{*}{\textbf{\begin{tabular}[c]{@{}c@{}}Victim \\ Models\end{tabular}}} & \multirow{2}{*}{\textbf{\begin{tabular}[c]{@{}c@{}}Attack \\ Goals\end{tabular}}} & \multirow{2}{*}{\textbf{\begin{tabular}[c]{@{}c@{}}Attack \\ Methods\end{tabular}}} & \multicolumn{2}{c|}{\textbf{Switzerland}} & \multicolumn{2}{c|}{\textbf{Chicago}} & \multicolumn{2}{c|}{\textbf{Joe Biden}} & \multicolumn{2}{c}{\textbf{Cornell University}} \\ 
\cmidrule(lr){4-5} \cmidrule(lr){6-7} \cmidrule(lr){8-9} \cmidrule(lr){10-11}
 & & & \,\,\,ACC &\,\,\,ASR& \,\,\,ACC & \,\,\,ASR & \,\,\,ACC & \,\,\,ASR& \,\,\,ACC & \,\,\,\,ASR \\ 
\midrule
\multirow{7}{*}{G-Retriever} & None & None   &  \multicolumn{1}{c|}{0.68} & \multicolumn{1}{c|}{-} &  \multicolumn{1}{c|}{0.73} & \multicolumn{1}{c|}{ -} & \multicolumn{1}{c|}{0.67}  & \multicolumn{1}{c|}{-}  & \multicolumn{1}{c|}{0.55}  &  \multicolumn{1}{c}{-} \\\cmidrule(lr){2-11}
&\multirow{3}{*}{DoS} & ACPI   &  \multicolumn{1}{c|}{0.68} & \multicolumn{1}{c|}{0.00} &  \multicolumn{1}{c|}{0.73} & \multicolumn{1}{c|}{ 0.00} & \multicolumn{1}{c|}{0.67}  & \multicolumn{1}{c|}{0.00}  & \multicolumn{1}{c|}{0.55}  &  \multicolumn{1}{c}{0.00} \\
 &                         & GCG   &  \multicolumn{1}{c|}{0.66} & \multicolumn{1}{c|}{0.00} & \multicolumn{1}{c|}{0.74}  & \multicolumn{1}{c|}{0.00}  & \multicolumn{1}{c|}{0.63}  & \multicolumn{1}{c|}{0.00} & \multicolumn{1}{c|}{0.50} & \multicolumn{1}{c}{0.03}  \\
 &                         & BadSKP    & \multicolumn{1}{c|}{ 0.66} &  \multicolumn{1}{c|}{\textbf{0.93}} &  \multicolumn{1}{c|}{0.71} &  \multicolumn{1}{c|}{\textbf{1.00}} & \multicolumn{1}{c|}{ 0.64}  & \multicolumn{1}{c|}{ \textbf{0.97}}   &  \multicolumn{1}{c|}{0.51} &    \multicolumn{1}{c}{\textbf{0.97}} \\
\cmidrule(lr){2-11}
&\multirow{3}{*}{IrA} & ACPI   &\multicolumn{1}{c|}{0.68}   &\multicolumn{1}{c|}{0.62 } &\multicolumn{1}{c|}{0.74}  &\multicolumn{1}{c|}{0.33} &\multicolumn{1}{c|}{0.67} &\multicolumn{1}{c|}{0.56 } &\multicolumn{1}{c|}{0.53}  & \multicolumn{1}{c}{0.56}\\
&                          & GCG   &\multicolumn{1}{c|}{0.65}   &\multicolumn{1}{c|}{0.62}  &\multicolumn{1}{c|}{0.72 }&\multicolumn{1}{c|}{0.51} &\multicolumn{1}{c|}{0.68}  & \multicolumn{1}{c|}{0.60} &\multicolumn{1}{c|}{0.52} &\multicolumn{1}{c}{0.46} \\
&                          & BadSKP  & \multicolumn{1}{c|}{0.68} &\multicolumn{1}{c|}{\textbf{0.88}  }&\multicolumn{1}{c|}{0.71} &\multicolumn{1}{c|}{\textbf{0.88} } &\multicolumn{1}{c|}{0.65} &\multicolumn{1}{c|}{\textbf{0.95}}&\multicolumn{1}{c|}{0.51}&\multicolumn{1}{c}{\textbf{1.00}}\\
\midrule
\multirow{7}{*}{GNP} & None & None   &\multicolumn{1}{c|}{0.57} &\multicolumn{1}{c|}{-} & \multicolumn{1}{c|}{0.60} &\multicolumn{1}{c|}{-}  &\multicolumn{1}{c|}{0.56}  & \multicolumn{1}{c|}{-} &\multicolumn{1}{c|}{0.55}  & \multicolumn{1}{c}{-} \\ \cmidrule(lr){2-11}
& \multirow{3}{*}{DoS} & ACPI   &\multicolumn{1}{c|}{0.57} &\multicolumn{1}{c|}{0.00} & \multicolumn{1}{c|}{0.60} &\multicolumn{1}{c|}{0.00}  &\multicolumn{1}{c|}{0.56}  & \multicolumn{1}{c|}{0.00} &\multicolumn{1}{c|}{0.55}  & \multicolumn{1}{c}{0.00} \\
 &                         & GCG   &\multicolumn{1}{c|}{0.56}  &\multicolumn{1}{c|}{0.13} &\multicolumn{1}{c|}{0.58}  &\multicolumn{1}{c|}{0.00}  &\multicolumn{1}{c|}{0.54}  &\multicolumn{1}{c|}{0.00}  &\multicolumn{1}{c|}{0.56} & \multicolumn{1}{c}{0.03}  \\
 &                         & BadSKP      & \multicolumn{1}{c|}{0.54} & \multicolumn{1}{c|}{\textbf{0.72}} & \multicolumn{1}{c|}{{0.55} }& \multicolumn{1}{c|}{\textbf{0.70} }&\multicolumn{1}{c|}{0.56} & \multicolumn{1}{c|}{\textbf{0.88}} &\multicolumn{1}{c|}{0.56}   &\multicolumn{1}{c}{\textbf{0.81}}  \\
\cmidrule(lr){2-11}
&\multirow{3}{*}{IrA} & ACPI   &\multicolumn{1}{c|}{0.53}   & \multicolumn{1}{c|}{0.79} & \multicolumn{1}{c|}{0.57}   & \multicolumn{1}{c|}{0.67}  &\multicolumn{1}{c|}{0.55} &\multicolumn{1}{c|}{0.77}  &\multicolumn{1}{c|}{0.56}  & \multicolumn{1}{c}{0.87}\\
&                          & GCG   &\multicolumn{1}{c|}{0.56}   &\multicolumn{1}{c|}{0.75}  &\multicolumn{1}{c|}{0.60} &\multicolumn{1}{c|}{0.80} &\multicolumn{1}{c|}{0.55}  &\multicolumn{1}{c|}{0.88}  &\multicolumn{1}{c|}{0.55} &\multicolumn{1}{c}{0.87} \\
&                          & BadSKP  & \multicolumn{1}{c|}{0.57} & \multicolumn{1}{c|}{\textbf{0.90}} & \multicolumn{1}{c|}{0.62} & \multicolumn{1}{c|}{\textbf{0.8}0}&\multicolumn{1}{c|}{0.56} &\multicolumn{1}{c|}{\textbf{0.91}}&\multicolumn{1}{c|}{0.56}&\multicolumn{1}{c}{\textbf{0.90}}\\
\bottomrule
\end{tabular}
\end{table*}

\subsubsection{Victim Systems}
We evaluate \textit{BadSKP} on two representative soft KG prompt systems, \textit{G-Retriever}~\cite{g-retriever} and \textit{GNP}~\cite{GNP}. To assess generalization across architectures, we consider four LLM backbones (\textit{LLaMA2-7B}~\cite{touvron2023llama}, \textit{LLaMA3-8B}~\cite{touvron2024llama3}, \textit{Mistral-8B}~\cite{jiang2023mistral}, and \textit{Qwen3-8B}~\cite{qwen3technicalreport}) and four GNN encoders (\textit{GAT}~\cite{velivckovic2018graph}, \textit{GCN}~\cite{kipf2017semi}, \textit{Graph Transformer}~\cite{DBLP:conf/nips/YunJKKK19}, and \textit{CGCNN}~\cite{xie2018crystal}). We also evaluate the attack under a perplexity-based filtering defense~\cite{DBLP:journals/corr/abs-2309-00614,DBLP:journals/corr/abs-2308-14132}.

\subsubsection{Evaluation Metrics}
We report two primary metrics.
\textit{Accuracy (ACC)} measures system utility and is computed as Hits@1 on clean, non-trigger queries.
\textit{Attack Success Rate (ASR)} measures attack effectiveness and is defined as the fraction of trigger-entity queries that elicit the attacker-specified response. Following the substring-matching convention in~\cite{DBLP:conf/iclr/HuangGXL024}, an attack is considered successful if the target string (or a refusal-style keyword, in the case of DoS) appears as a substring of the generated output. We adopt this convention for comparability with prior work.

\begin{table*}[htbp]
\centering
\caption{ACC and ASR of different attack methods against G-Retriever with a trojaned soft prompt module (No Defense).}
\label{tab:pretrained_gnn_results_no_defense}
\begin{tabular}{@{}cc *{8}{|p{0.80cm}}@{}}
\toprule
& & \multicolumn{4}{c|}{\textbf{CWQ}} & \multicolumn{4}{c}{\textbf{WEBQSP}} \\ 
\cmidrule(lr){3-6} \cmidrule(lr){7-10}
\multirow{2}{*}{\textbf{\begin{tabular}[c]{@{}c@{}}Attack \\ Goals\end{tabular}}} & \multirow{2}{*}{\textbf{\begin{tabular}[c]{@{}c@{}}Attack \\ Methods\end{tabular}}} & \multicolumn{2}{c|}{\textbf{Switzerland}} & \multicolumn{2}{c|}{\textbf{Chicago}} & \multicolumn{2}{c|}{\textbf{Joe Biden}} & \multicolumn{2}{c}{\textbf{Cornell University}} \\ 
\cmidrule(lr){3-4} \cmidrule(lr){5-6} \cmidrule(lr){7-8} \cmidrule(lr){9-10}
 & & \,\,\,ACC & \,\,\,ASR & \,\,\,ACC & \,\,\,ASR & \,\,\,ACC & \,\,\,ASR & \,\,\,ACC & \,\,\,\,ASR \\ 
\midrule
\multirow{4}{*}{DoS} & None & \multicolumn{1}{c|}{0.61} & \multicolumn{1}{c|}{-} & \multicolumn{1}{c|}{0.66} & \multicolumn{1}{c|}{-} & \multicolumn{1}{c|}{0.57} & \multicolumn{1}{c|}{-} & \multicolumn{1}{c|}{0.56} & \multicolumn{1}{c}{-} \\ \cmidrule(lr){2-10}
& ACPI  & \multicolumn{1}{c|}{0.61} & \multicolumn{1}{c|}{0.00} & \multicolumn{1}{c|}{0.66} & \multicolumn{1}{c|}{0.00} & \multicolumn{1}{c|}{0.57} & \multicolumn{1}{c|}{0.00} & \multicolumn{1}{c|}{0.56} & \multicolumn{1}{c}{0.00} \\
& GCG   & \multicolumn{1}{c|}{0.63} & \multicolumn{1}{c|}{0.00} & \multicolumn{1}{c|}{0.62} & \multicolumn{1}{c|}{0.00} & \multicolumn{1}{c|}{0.60} & \multicolumn{1}{c|}{0.00} & \multicolumn{1}{c|}{0.46} & \multicolumn{1}{c}{0.00} \\
& BadSKP & \multicolumn{1}{c|}{0.64} & \multicolumn{1}{c|}{\textbf{1.00}} & \multicolumn{1}{c|}{0.66} & \multicolumn{1}{c|}{\textbf{1.00}} & \multicolumn{1}{c|}{0.64} & \multicolumn{1}{c|}{\textbf{0.97}} & \multicolumn{1}{c|}{0.51} & \multicolumn{1}{c}{\textbf{0.87}} \\ 
\cmidrule(lr){1-10}
\multirow{4}{*}{IrA} & None & \multicolumn{1}{c|}{0.43} & \multicolumn{1}{c|}{-} & \multicolumn{1}{c|}{0.67} & \multicolumn{1}{c|}{-} & \multicolumn{1}{c|}{0.57} & \multicolumn{1}{c|}{-} & \multicolumn{1}{c|}{0.53} & \multicolumn{1}{c}{-} \\ \cmidrule(lr){2-10}
& ACPI  & \multicolumn{1}{c|}{0.43} & \multicolumn{1}{c|}{0.58} & \multicolumn{1}{c|}{0.67} & \multicolumn{1}{c|}{0.59} & \multicolumn{1}{c|}{0.57} & \multicolumn{1}{c|}{0.53} & \multicolumn{1}{c|}{0.44} & \multicolumn{1}{c}{0.40} \\
& GCG   & \multicolumn{1}{c|}{0.40} & \multicolumn{1}{c|}{0.69} & \multicolumn{1}{c|}{0.64} & \multicolumn{1}{c|}{0.51} & \multicolumn{1}{c|}{0.57} & \multicolumn{1}{c|}{0.62} & \multicolumn{1}{c|}{0.53} & \multicolumn{1}{c}{0.28} \\
& BadSKP & \multicolumn{1}{c|}{0.42} & \multicolumn{1}{c|}{\textbf{1.00}} & \multicolumn{1}{c|}{0.64} & \multicolumn{1}{c|}{\textbf{1.00}} & \multicolumn{1}{c|}{0.62} & \multicolumn{1}{c|}{\textbf{1.00}} & \multicolumn{1}{c|}{0.48} & \multicolumn{1}{c}{\textbf{1.00}} \\
\bottomrule
\end{tabular}
\end{table*}

\subsection{Experimental Results}

We evaluate \textit{BadSKP} against two text-only baselines, \textit{ACPI} and \textit{GCG}, on two soft-prompt-based KG-enhanced LLMs under both attack objectives (DoS and IrA) across four trigger entities. Table~\ref{tab:fixed_gnn_results_no_defense} reports ASR and ACC for \textit{G-Retriever} and \textit{GNP} with a frozen soft prompt module and a LLaMA~2-7B backbone, while Table~\ref{tab:pretrained_gnn_results_no_defense} reports the trojaned setting. Higher ASR indicates stronger attack effectiveness, whereas higher ACC indicates better benign QA utility.


\subsubsection{Attack Performance in the Frozen Setting}
As shown in Table~\ref{tab:fixed_gnn_results_no_defense}, \textit{BadSKP} achieves strong attack effectiveness in the frozen setting. Under the DoS objective, its ASR consistently exceeds 0.70, whereas both text-only baselines remain below 0.15 in nearly all cases. Under the IrA objective, \textit{BadSKP} also outperforms the baselines across all trigger entities and victim models, while maintaining ACC close to that of the clean model.
This gap highlights a key limitation of text-only attacks in soft KG prompting systems. \textit{ACPI} and \textit{GCG} perturb node attributes through adversarial textual content but do not manipulate the graph-derived soft prompt, and thus fail to overcome the semantic anchoring effect identified in Section~\ref{subsec:mechanism}. By contrast, \textit{BadSKP} perturbs both the textual and graph channels, allowing the fused latent representation to shift toward attacker-specified semantics.



\subsubsection{Attack Performance in the Trojaned Setting}
Table~\ref{tab:pretrained_gnn_results_no_defense} shows that \textit{BadSKP} remains effective when the soft prompt module is trojaned and subsequently fine-tuned on clean data. Under the DoS objective, \textit{BadSKP} achieves ASR above 0.87 across all trigger entities, while both text-only baselines remain at or near zero. Under the IrA objective, \textit{BadSKP} reaches ASR of 1.00 in three out of four settings, substantially outperforming the baselines by margins exceeding 0.30.
This persistence is supported by two design choices. First, the adversarial suffixes optimized in Steps~1--3 provide a strong backdoor signal at initialization. Second, the gradient-alignment loss $\mathcal{L}_{\mathrm{aln}}$
encourages backdoor-preserving updates to remain compatible with the main task, reducing the extent to which clean fine-tuning erases the implanted behavior.



\subsubsection{Generalization across GNN and LLM Architectures}
We further examine whether the attack remains effective across different GNN modules and LLM backbones. Across diverse GNN encoders (GCN, GAT, CGCNN, and Graph Transformer) and LLM backbones (LLaMA2-7B, LLaMA3-8B, Mistral-8B, and Qwen3-8B), \textit{BadSKP} consistently attains high ASR under both DoS and IrA objectives, while remaining substantially stronger than the text-only baselines.
\begin{table}[]
\centering
\caption{ACC and ASR of attack methods against G-Retriever with frozen soft KG prompt module using different base GNN models.}
\label{tab:diff_gnn}
\resizebox{\columnwidth}{!}{
\begin{tabular}{cc|cc|cc|cc|cc}
\toprule
\multirow{2}{*}{\textbf{\begin{tabular}[c]{@{}c@{}}Attack \\ Goals\end{tabular}}} & \multirow{2}{*}{\textbf{\begin{tabular}[c]{@{}c@{}}Attack\\ Methods\end{tabular}}} & \multicolumn{2}{c|}{\textbf{GCN}}         & \multicolumn{2}{c|}{\textbf{GAT}}         & \multicolumn{2}{c|}{\textbf{CGCNN}} & \multicolumn{2}{c}{\textbf{GT}}           \\ \cmidrule{3-10} 
                                                                                  &                                                                                    & \multicolumn{1}{c|}{ACC}  & ASR           & \multicolumn{1}{c|}{ACC}  & ASR           & \multicolumn{1}{c|}{ACC}    & ASR   & \multicolumn{1}{c|}{ACC}  & ASR           \\  \midrule
\multirow{3}{*}{DoS}                                                              & ACPI                                                                               & \multicolumn{1}{c|}{0.58} & 0.22          & \multicolumn{1}{c|}{0.73} & 0.00          & \multicolumn{1}{c|}{0.74}   & 0.62  & \multicolumn{1}{c|}{0.75} & 0.44          \\ \cmidrule{2-10} 
                                                                                  & GCG                                                                                & \multicolumn{1}{c|}{0.63} & 0.00          & \multicolumn{1}{c|}{0.74} & 0.00          & \multicolumn{1}{c|}{0.65}   & 0.00  & \multicolumn{1}{c|}{0.71} & 0.00          \\ \cmidrule{2-10} 
                                                                                  & BadSKP                                                                             & \multicolumn{1}{c|}{0.66} & \textbf{1.00} & \multicolumn{1}{c|}{0.71} & \textbf{1.00} & \multicolumn{1}{c|}{0.74}   & \textbf{0.73}  & \multicolumn{1}{c|}{0.74} & \textbf{0.70} \\  \midrule
\multirow{3}{*}{IrA}                                                              & ACPI                                                                               & \multicolumn{1}{c|}{0.58} & 0.33          & \multicolumn{1}{c|}{0.74} & 0.33          & \multicolumn{1}{c|}{0.62}   & 0.74  & \multicolumn{1}{c|}{0.75} & \textbf{0.81} \\ \cmidrule{2-10} 
                                                                                  & GCG                                                                                & \multicolumn{1}{c|}{0.66} & 0.44          & \multicolumn{1}{c|}{0.72} & 0.51          & \multicolumn{1}{c|}{0.66}   & 0.44  & \multicolumn{1}{c|}{0.70} & 0.77          \\ \cmidrule{2-10} 
                                                                                  & BadSKP                                                                             & \multicolumn{1}{c|}{0.65}     & \textbf{0.92}     & \multicolumn{1}{c|}{0.71} & \textbf{0.88} & \multicolumn{1}{c|}{0.70}       &  \textbf{0.70}     & \multicolumn{1}{c|}{0.72} & 0.70          \\ \bottomrule
\end{tabular}
}
\end{table}

\begin{table}[]
\centering
\caption{ACC and ASR of attack methods for the DoS objective against {G-Retriever} with a frozen soft prompt module under different LLM backbones.}
\label{tab:diff_llm}
\resizebox{\columnwidth}{!}{
\begin{tabular}{c|cc|cc|cc|cc}
\toprule
\multirow{2}{*}{\textbf{\begin{tabular}[c]{@{}c@{}}Attack \\ Methods\end{tabular}}} & \multicolumn{2}{c|}{\textbf{LLAMA2}} & \multicolumn{2}{c|}{\textbf{LLAMA3}} & \multicolumn{2}{c|}{\textbf{Ministral}} & \multicolumn{2}{c}{\textbf{Qwen3}} \\ \cmidrule{2-9} 
                                                                                    & \multicolumn{1}{c|}{ACC}    & ASR    & \multicolumn{1}{c|}{ACC}    & ASR    & \multicolumn{1}{c|}{ACC}      & ASR     & \multicolumn{1}{c|}{ACC}   & ASR  \\ \midrule
ACPI                                                                               & \multicolumn{1}{c|}{0.72}   & 0.00   & \multicolumn{1}{c|}{0.66}   & 0.00   & \multicolumn{1}{c|}{0.65}         &  0.00       & \multicolumn{1}{c|}{0.68}      &   0.00   \\ \midrule
GCG                                                                                 & \multicolumn{1}{c|}{0.74}   & 0.00   & \multicolumn{1}{c|}{0.72}   & 0.55   & \multicolumn{1}{c|}{0.65}         & 0.00        & \multicolumn{1}{c|}{0.68}  & 0.00 \\ \midrule
BadSKP                                                                             & \multicolumn{1}{c|}{0.71}   & \multicolumn{1}{c|}{\textbf{1.00}}  & \multicolumn{1}{c|}{0.72}   & \multicolumn{1}{c|}{\textbf{0.81}}   & \multicolumn{1}{c|}{0.65}         & \multicolumn{1}{c|}{\textbf{0.80} }      & \multicolumn{1}{c|}{0.67}  & \textbf{1.00} \\ \bottomrule
\end{tabular}
}
\end{table}


Table~\ref{tab:diff_gnn} reports the performance of \textit{BadSKP} against \textit{G-Retriever} with four GNN encoders (GCN, GAT, CGCNN, and Graph Transformer) in the frozen setting. Across all encoders, \textit{BadSKP} maintains ASR above 0.70 under both DoS and IrA objectives, suggesting that the attack is not tied to a particular graph encoder architecture.
Table~\ref{tab:diff_llm} presents the results under the DoS objective with four LLM backbones (\textit{LLaMA2-7B}, \textit{LLaMA3-8B}, \textit{Mistral-8B}, and \textit{Qwen3-8B}). \textit{BadSKP} achieves ASR above 0.80 across all backbones, whereas the text-only baselines remain below 0.55. The consistent performance across models with different pretraining data, tokenization schemes, and alignment characteristics suggests that \textit{BadSKP} exploits a structural weakness in graph--text fusion rather than model-specific decoding behavior.

\subsubsection{Effect of Adversarial Suffix Length}
\begin{figure}[htb]
    \centering
    \includegraphics[height=4cm,width=9cm]{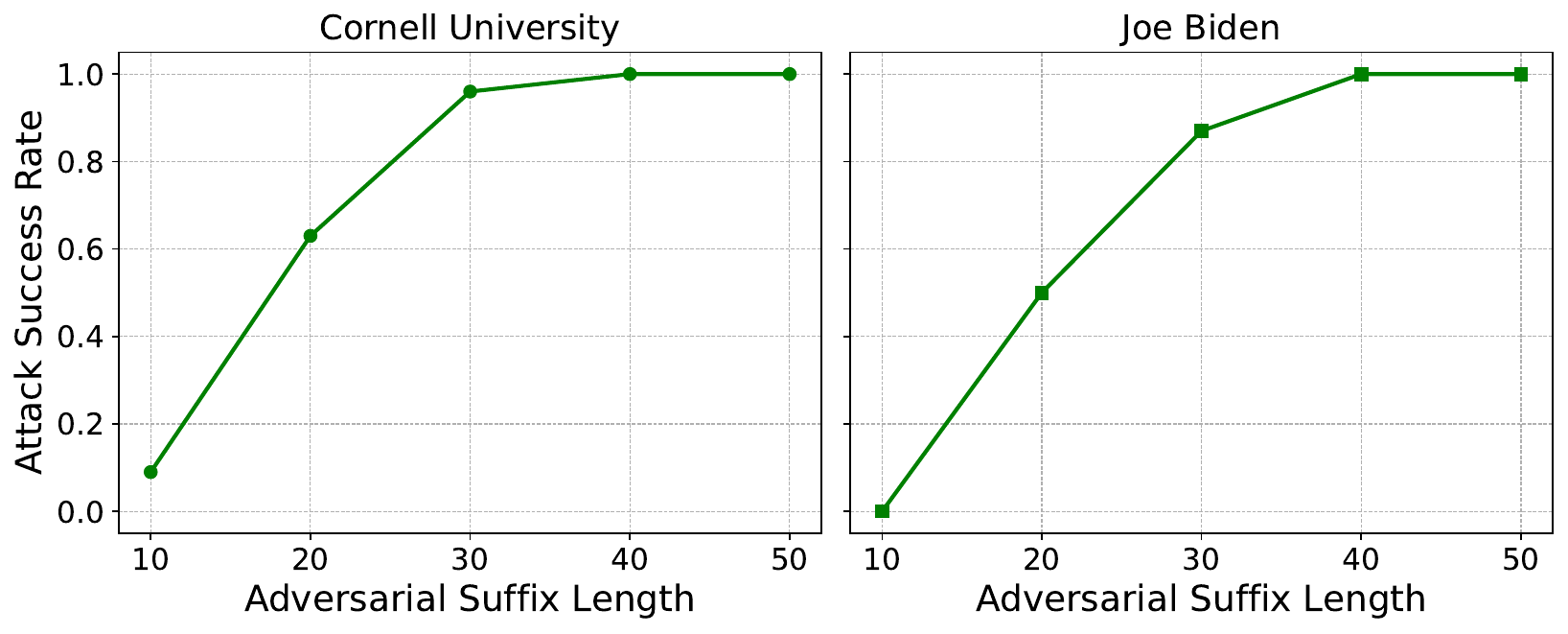}
    \caption{Effect of adversarial suffix length $M$ on the ASR of \textit{BadSKP} with the DoS objective.}
    \label{fig:tokennum}
\end{figure}

Figure~\ref{fig:tokennum} shows the effect of adversarial suffix length $M \in \{10,20,30,40,50\}$ on the ASR of \textit{BadSKP} against \textit{G-Retriever} on WebQSP, using \textit{Cornell University} as the trigger entity. As $M$ increases from 10 to 30, ASR rises substantially, indicating that longer suffixes allow a more accurate approximation of the optimized target embedding. Beyond $M=30$, the gain becomes marginal, suggesting diminishing returns from further increasing the suffix length.

\subsubsection{Effect of Number of Poisoned Nodes}
\begin{figure}[htb]
    \centering
    \includegraphics[height=4cm,width=9cm]{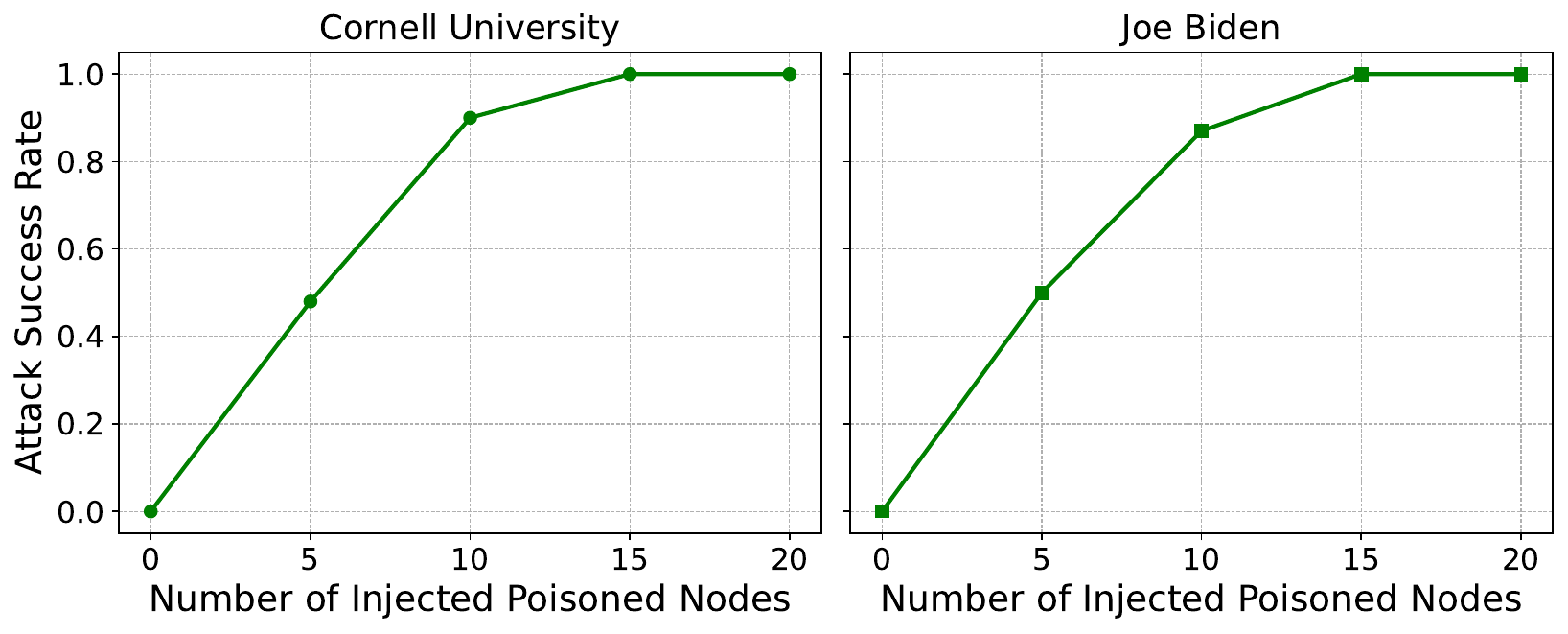}
    \caption{Effect of the number of poisoned nodes $K$ on the ASR of \textit{BadSKP} with the DoS objective.}
    \label{fig:poisonednode}
\end{figure}

Figure~\ref{fig:poisonednode} shows the effect of the number of injected poisoned nodes $K \in \{0,5,10,15,20\}$ on the ASR of \textit{BadSKP}. ASR increases monotonically with $K$, indicating that aggregating adversarial signals over more poisoned nodes progressively weakens the semantic anchoring induced by the soft KG prompt. Each additional poisoned node contributes to shifting the global graph representation toward the adversarial target semantics, thereby making backdoor activation more reliable.

\begin{table*}[htb]
\centering
\caption{ACC and ASR of different attack methods against victim systems with a frozen soft prompt module (Perplexity Defense).}
\label{tab:fixed_gnn_results_defense}
\begin{tabular}{@{}ccc *{8}{|p{0.80cm}}@{}}
\toprule
&& & \multicolumn{4}{c|}{\textbf{CWQ}} & \multicolumn{4}{c}{\textbf{WEBQSP}} \\ 
\cmidrule(lr){4-7} \cmidrule(lr){8-11}
\multirow{2}{*}{\textbf{\begin{tabular}[c]{@{}c@{}}Victim \\ Models\end{tabular}}} & \multirow{2}{*}{\textbf{\begin{tabular}[c]{@{}c@{}}Attack \\ Goals\end{tabular}}} & \multirow{2}{*}{\textbf{\begin{tabular}[c]{@{}c@{}}Attack \\ Methods\end{tabular}}} & \multicolumn{2}{c|}{\textbf{Switzerland}} & \multicolumn{2}{c|}{\textbf{Chicago}} & \multicolumn{2}{c|}{\textbf{Joe Biden}} & \multicolumn{2}{c}{\textbf{Cornell University}} \\ 
\cmidrule(lr){4-5} \cmidrule(lr){6-7} \cmidrule(lr){8-9} \cmidrule(lr){10-11}
 & & & \,\,\,ACC &\,\,\,ASR& \,\,\,ACC & \,\,\,ASR & \,\,\,ACC & \,\,\,ASR& \,\,\,ACC & \,\,\,\,ASR \\ 
\midrule
\multirow{6}{*}{G-Retriever} &\multirow{3}{*}{DoS} & ACPI   & \multicolumn{1}{c|}{0.37} & \multicolumn{1}{c|}{0.00 } & \multicolumn{1}{c|}{0.44}  & \multicolumn{1}{c|}{0.00}  & \multicolumn{1}{c|}{0.37} & \multicolumn{1}{c|}{0.00} & \multicolumn{1}{c|}{0.40} & \multicolumn{1}{c}{0.00} \\
&                          & GCG    &\multicolumn{1}{c|}{ 0.41}   & \multicolumn{1}{c|}{0.00}  & \multicolumn{1}{c|}{0.36} & \multicolumn{1}{c|}{0.00}  & \multicolumn{1}{c|}{0.20} & \multicolumn{1}{c|}{0.00} & \multicolumn{1}{c|}{0.24} & \multicolumn{1}{c}{0.00 } \\
&                          &  BadSKP     & \multicolumn{1}{c|}{{0.40}} &\multicolumn{1}{c|}{\textbf{0.79}}  &\multicolumn{1}{c|}{{0.45}} &\multicolumn{1}{c|}{\textbf{0.92}}  & \multicolumn{1}{c|}{0.36}  & \multicolumn{1}{c|}{\textbf{0.88}} & \multicolumn{1}{c|}{0.20} & \multicolumn{1}{c}{\textbf{0.97}} \\
\cmidrule(lr){2-11}
&\multirow{3}{*}{IrA} & ACPI   &\multicolumn{1}{c|}{0.42}  &\multicolumn{1}{c|}{0.55}  &\multicolumn{1}{c|}{0.46}  &\multicolumn{1}{c|}{0.37 }  &\multicolumn{1}{c|}{0.40 }&\multicolumn{1}{c|}{0.66 } &\multicolumn{1}{c|}{0.40} &\multicolumn{1}{c}{0.53 } \\
 &                         & GCG   &\multicolumn{1}{c|}{0.38}  &\multicolumn{1}{c|}{0.69} &\multicolumn{1}{c|}{0.43}&\multicolumn{1}{c|}{0.37} & \multicolumn{1}{c|}{0.25}& \multicolumn{1}{c|}{0.55}&\multicolumn{1}{c|}{0.20} &\multicolumn{1}{c}{0.46}  \\
&                          &  BadSKP     &\multicolumn{1}{c|}{0.40}  &\multicolumn{1}{c|}{\textbf{0.90}}& \multicolumn{1}{c|}{0.43}&  \multicolumn{1}{c|}{\textbf{1.00}}& \multicolumn{1}{c|}{0.37} & \multicolumn{1}{c|}{\textbf{0.91}}&\multicolumn{1}{c}{0.30}  &\multicolumn{1}{c}{\textbf{0.90}}  \\
\midrule
\multirow{6}{*}{GNP} & \multirow{3}{*}{DoS} & ACPI   &\multicolumn{1}{c|}{0.40} & \multicolumn{1}{c|}{0.00} &\multicolumn{1}{c|}{0.26}  & \multicolumn{1}{c|}{0.00}  &\multicolumn{1}{c|}{0.33}  & \multicolumn{1}{c|}{0.00} &\multicolumn{1}{c|}{0.18} &\multicolumn{1}{c}{0.00} \\
&                          & GCG    &\multicolumn{1}{c|}{0.38}   &\multicolumn{1}{c|}{0.02 } &\multicolumn{1}{c|}{0.33 }&\multicolumn{1}{c|}{0.00 } &\multicolumn{1}{c|}{0.20} &\multicolumn{1}{c|}{0.00}&\multicolumn{1}{c|}{0.20} & \multicolumn{1}{c}{0.03} \\
&                          & BadSKP  & \multicolumn{1}{c|}{0.35 }& \multicolumn{1}{c|}{\textbf{0.75} }&\multicolumn{1}{c|}{{0.30} } & \multicolumn{1}{c|}{\textbf{0.70} } &\multicolumn{1}{c|}{0.30}  &\multicolumn{1}{c|}{\textbf{0.88}} &\multicolumn{1}{c|}{0.21 }&\multicolumn{1}{c}{\textbf{0.75}} \\
\cmidrule(lr){2-11}
&\multirow{3}{*}{IrA} & ACPI   & \multicolumn{1}{c|}{0.40}  & \multicolumn{1}{c|}{0.79}  & \multicolumn{1}{c|}{0.28} & \multicolumn{1}{c|}{{0.74}}  & \multicolumn{1}{c|}{0.23} & \multicolumn{1}{c|}{0.73}  & \multicolumn{1}{c|}{0.20}& \multicolumn{1}{c}{0.87}  \\
 &                         & GCG   & \multicolumn{1}{c|}{0.35} &\multicolumn{1}{c|}{0.88} &\multicolumn{1}{c|}{0.34} &\multicolumn{1}{c|}{0.67} &\multicolumn{1}{c|}{0.20} &\multicolumn{1}{c|}{\textbf{0.83}} &\multicolumn{1}{c|}{0.21} & \multicolumn{1}{c}{\textbf{0.87}} \\
&                          & BadSKP     & \multicolumn{1}{c|}{0.30} &\multicolumn{1}{c|}{\textbf{0.97}} &\multicolumn{1}{c|}{0.31} & \multicolumn{1}{c|}{\textbf{0.74}} &\multicolumn{1}{c|}{0.21} &\multicolumn{1}{c|}{0.73} &\multicolumn{1}{c|}{0.20}  &\multicolumn{1}{c}{ 0.84} \\
\bottomrule
\end{tabular}
\end{table*}

\begin{table*}[htbp]
\centering
\caption{ACC and ASR of different attack methods against G-Retriever with a trojaned soft prompt module (Perplexity Defense).}
\label{tab:pretrained_gnn_results_perplexity}
\begin{tabular}{@{}cc *{8}{|p{0.80cm}}@{}}
\toprule
& & \multicolumn{4}{c|}{\textbf{CWQ}} & \multicolumn{4}{c}{\textbf{WEBQSP}} \\ 
\cmidrule(lr){3-6} \cmidrule(lr){7-10}
\multirow{2}{*}{\textbf{\begin{tabular}[c]{@{}c@{}}Attack \\ Goals\end{tabular}}} & \multirow{2}{*}{\textbf{\begin{tabular}[c]{@{}c@{}}Attack \\ Methods\end{tabular}}} & \multicolumn{2}{c|}{\textbf{Switzerland}} & \multicolumn{2}{c|}{\textbf{Chicago}} & \multicolumn{2}{c|}{\textbf{Joe Biden}} & \multicolumn{2}{c}{\textbf{Cornell University}} \\ 
\cmidrule(lr){3-4} \cmidrule(lr){5-6} \cmidrule(lr){7-8} \cmidrule(lr){9-10}
 & & \,\,\,ACC & \,\,\,ASR & \,\,\,ACC & \,\,\,ASR & \,\,\,ACC & \,\,\,ASR & \,\,\,ACC & \,\,\,\,ASR \\ 
\midrule
\multirow{3}{*}{DoS} & ACPI   & \multicolumn{1}{c|}{0.64} & \multicolumn{1}{c|}{0.00} & \multicolumn{1}{c|}{0.64} & \multicolumn{1}{c|}{0.00} & \multicolumn{1}{c|}{0.57} & \multicolumn{1}{c|}{0.00} & \multicolumn{1}{c|}{0.45} & \multicolumn{1}{c}{0.00} \\
& GCG   & \multicolumn{1}{c|}{0.58} & \multicolumn{1}{c|}{0.00} & \multicolumn{1}{c|}{0.64} & \multicolumn{1}{c|}{0.00} & \multicolumn{1}{c|}{0.60} & \multicolumn{1}{c|}{0.00} & \multicolumn{1}{c|}{0.55} & \multicolumn{1}{c}{0.00} \\
& BadSKP & \multicolumn{1}{c|}{0.59} & \multicolumn{1}{c|}{\textbf{1.00}} & \multicolumn{1}{c|}{0.64} & \multicolumn{1}{c|}{\textbf{1.00}} & \multicolumn{1}{c|}{0.64} & \multicolumn{1}{c|}{\textbf{0.71}} & \multicolumn{1}{c|}{0.55} & \multicolumn{1}{c}{\textbf{0.96}} \\ 
\cmidrule(lr){1-10}
\multirow{3}{*}{IrA} & ACPI   & \multicolumn{1}{c|}{0.42} & \multicolumn{1}{c|}{0.60} & \multicolumn{1}{c|}{0.63} & \multicolumn{1}{c|}{0.55} & \multicolumn{1}{c|}{0.56} & \multicolumn{1}{c|}{0.53} & \multicolumn{1}{c|}{0.52} & \multicolumn{1}{c}{0.28} \\
& GCG    & \multicolumn{1}{c|}{0.40} & \multicolumn{1}{c|}{0.00} & \multicolumn{1}{c|}{0.64} & \multicolumn{1}{c|}{0.33} & \multicolumn{1}{c|}{0.55} & \multicolumn{1}{c|}{0.40} & \multicolumn{1}{c|}{0.45} & \multicolumn{1}{c}{0.28} \\
& BadSKP & \multicolumn{1}{c|}{0.40} & \multicolumn{1}{c|}{\textbf{1.00}} & \multicolumn{1}{c|}{0.60} & \multicolumn{1}{c|}{\textbf{1.00}} & \multicolumn{1}{c|}{0.55} & \multicolumn{1}{c|}{\textbf{0.86}} & \multicolumn{1}{c|}{0.42} & \multicolumn{1}{c}{\textbf{1.00}} \\
\bottomrule
\end{tabular}
\end{table*}
\section{Potential Defense }
We first evaluate perplexity-based filtering and then discuss a defense perspective motivated by semantic anchoring.

\noindent\textbf{Perplexity-based Filtering.}
Text-based defenses such as perplexity filtering~\cite{DBLP:journals/corr/abs-2309-00614} and paraphrasing~\cite{DBLP:conf/uss/LiuJGJG24} are primarily designed for attacks on unstructured textual inputs. Their applicability to KG-enhanced LLMs is limited, because the retrieved evidence is organized as structured subgraphs where node attributes and graph topology jointly determine the soft prompt representation. Aggressive text rewriting may distort benign node attributes and compromise the semantic integrity of the retrieved context. 

To assess whether lightweight text-side sanitization can mitigate our attack, we evaluate a perplexity-based defense that filters node attributes whose perplexity exceeds a fixed threshold. As shown in Tables~\ref{tab:fixed_gnn_results_defense}--\ref{tab:pretrained_gnn_results_perplexity}, \textit{BadSKP} remains effective under this defense, particularly in the DoS setting. A plausible reason is that the adversarial suffixes in \textit{BadSKP} are synthesized through a language-model-guided search in Step~3, which constrains candidate tokens to locally fluent continuations and thus makes the injected content less likely to be removed by perplexity-based filtering. We also observe a noticeable drop in ACC after filtering is enabled, because the defense may mistakenly remove benign node attributes that are short, structured, or context-dependent rather than fully natural sentences. 

\noindent\textbf{Our Anchoring-strength Monitoring.}
The above findings suggest that defending soft-prompt-based KG-enhanced LLMs requires going beyond surface-level text filtering and instead monitoring whether the graph-induced soft prompt preserves the expected query-consistent semantic anchor during generation.
Our analysis in Section~\ref{subsec:mechanism} shows that, under benign inputs, graph-derived soft prompts consistently strengthen the association between the generation-driving hidden state $\mathbf{h}_{\mathrm{last}}^{(L)}$ and query-relevant tokens, while suppressing the influence of adversarial tokens. Motivated by this observation, we design a lightweight runtime defense that measures the anchoring strength of the current input. Specifically, we compute the cosine similarity between $\mathbf{h}_{\mathrm{last}}^{(L)}$ and the hidden states of query tokens, and use the maximum similarity score as an indicator of whether the generation remains anchored to the original query semantics. A detection threshold is calibrated on clean validation samples, where this score is typically high. Inputs whose anchoring-strength score falls below the threshold are flagged as suspicious, since such a drop suggests that the original query-consistent anchor may have been weakened or redirected by adversarial graph perturbations.
Experimental results show that this defense effectively detects poisoned inputs and reduces the ASR of \textit{BadSKP}, while preserving most benign-task performance. These results suggest that semantic anchoring is not only a useful mechanism for interpreting the robustness of soft KG prompts, but also a practical signal for runtime defense.

\section{Conclusion}
This work provides a systematic security analysis of soft-prompt-based KG-enhanced LLMs under backdoor attacks. We show that graph-derived soft prompts induce a semantic anchoring effect that stabilizes generation by constraining it to a query-aligned latent region, rendering text-only perturbations largely ineffective. Building on this insight, we propose \textit{BadSKP}, a backdoor attack that jointly manipulates graph structure and textual attributes to disrupt the anchoring constraint and hijack the latent semantic space. Our results demonstrate that the intrinsic robustness of soft prompts is not a security guarantee: once semantic anchoring is disrupted, model generation can be reliably misled. These findings highlight the graph-to-prompt interface as a critical security boundary that warrants dedicated defenses beyond conventional text-level sanitization.

\bibliographystyle{IEEEtran}
\bibliography{ref.bib}

\begin{IEEEbiography}[{\includegraphics[width=0.8in,height=1.25in,clip,keepaspectratio]{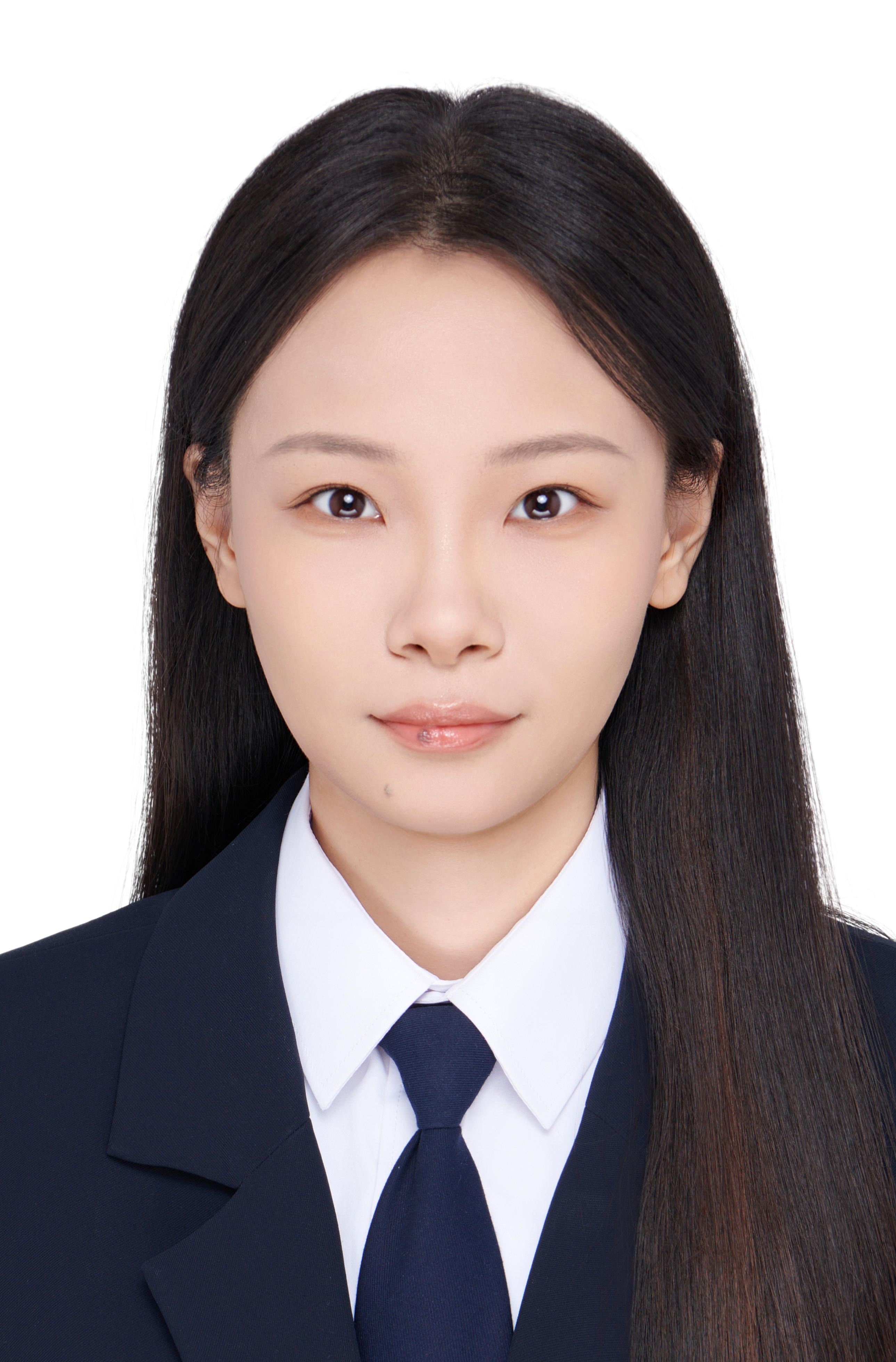}}]{Xiaoting Lyu} is the assistant professor with Ministry of Education Key Lab for Intelligent Networks and Network Security, aka MOE KLINNS Lab, Xi'an Jiaotong University, China. She received the Ph.D. degree from Beijing Jiaotong University in 2025. 
\end{IEEEbiography}

\begin{IEEEbiography}[{\includegraphics[width=0.8in,height=1.25in,clip,keepaspectratio]{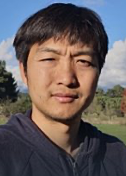}}]{Yufei Han} is a senior researcher at INRIA France. His research interests encompass trustworthy and interpretable AI technologies for cybersecurity applications, as well as the adversarial robustness of AI systems. He has published over 50 research papers on top-tiered venues on AI and cybersecurity, like IEEE S\&P, CCS, KDD, NDSS and AAAI. He regularly serves as program committees and peer reviews in these venues.
\end{IEEEbiography}

\begin{IEEEbiography}[{\includegraphics[width=0.8in,height=1.25in,clip,keepaspectratio]{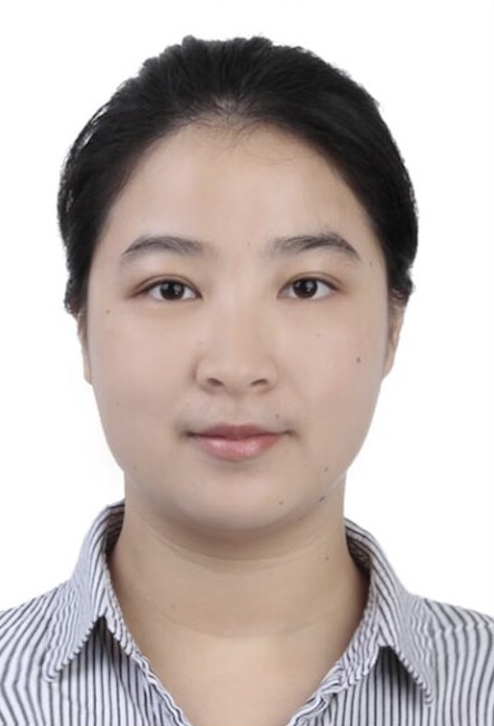}}]{Hangwei Qian} is currently a Scientist at Centre for Frontier AI Research (CFAR), A*STAR Research Entities, Singapore. Previously she was awarded the prestigeous Wallenberg-NTU Presidential Postdoctoral Fellowship between 2020 and 2022. She obtained her Ph.D. in computer science and engineering at Nanyang Technological University (NTU), Singapore in 2020. She received the B.Eng. from the University of Science and Technology of China (USTC) in 2015. Her research interests lie in Trustworthy AI, Transfer learning, and AI for Science. She has published top-tier papers on KDD, IJCAI, AAAI and AIJ over the years.
\end{IEEEbiography}

\begin{IEEEbiography}[{\includegraphics[width=0.8in,height=1.25in,clip,keepaspectratio]{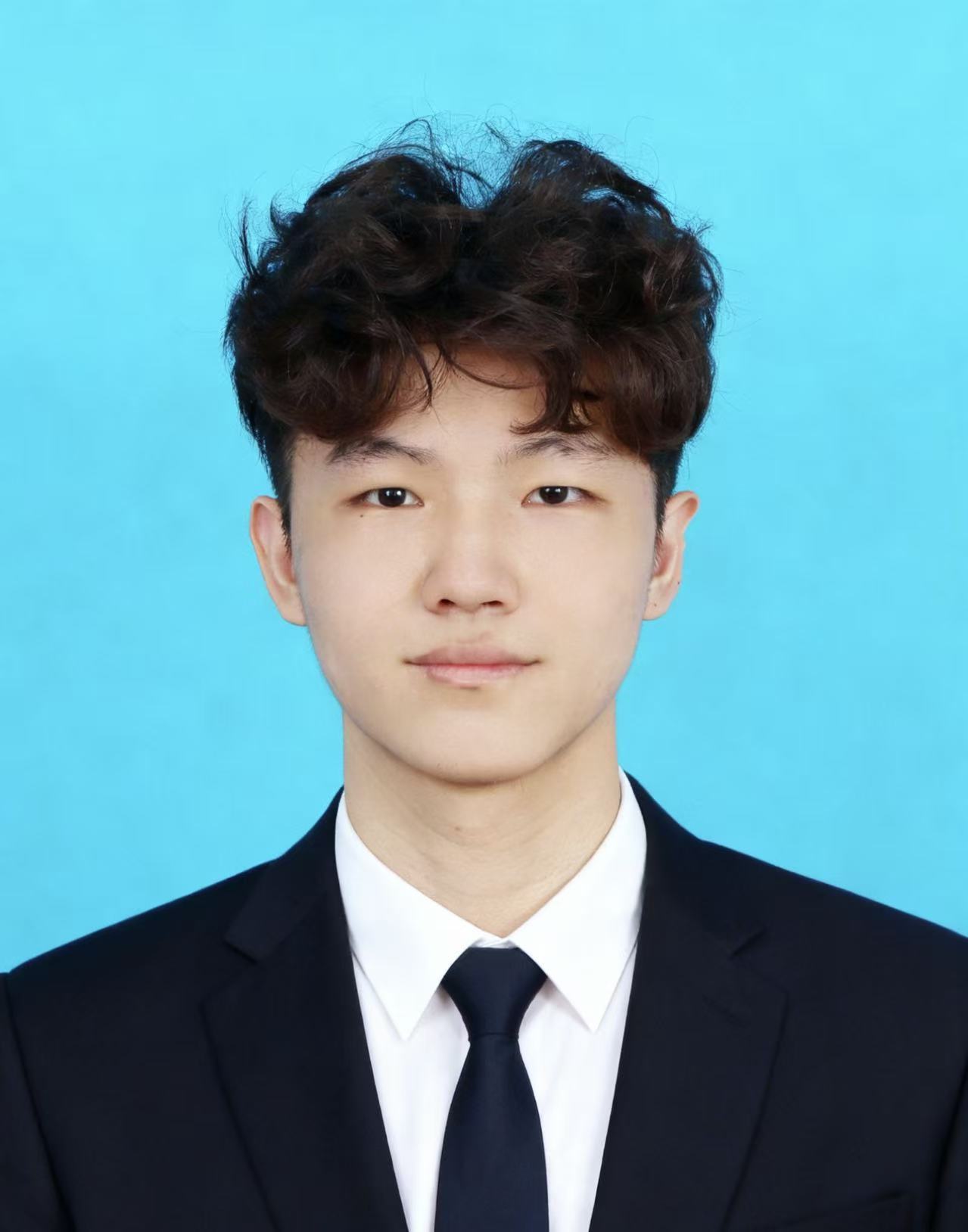}}]{Haoyuan Yu} is currently pursuing the bachelor's degree with Beijing Jiaotong University, Beijing, China, since 2022.
\end{IEEEbiography}

\begin{IEEEbiography}[{\includegraphics[width=0.8in,height=1.25in,clip,keepaspectratio]{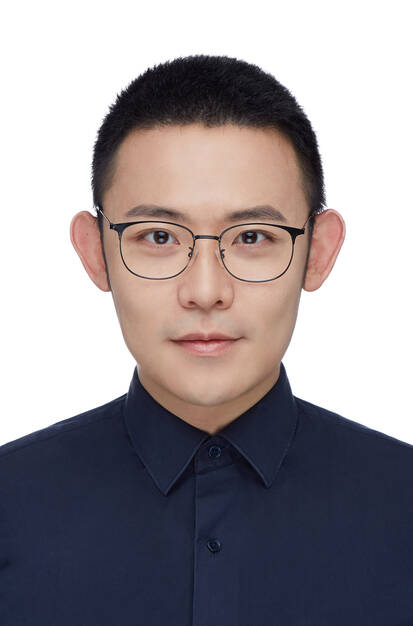}}]{Xiang Ao} is a Professor in the State Key Laboratory of AI Safety, Institute of Computing Technology, Chinese Academy of Sciences(ICT, CAS). Before joining ICT, he received Ph.D. degree in Computer Science from the Institute of Computing Technology, CAS in 2015 and B.S. degree in Computer Science from Zhejiang University in 2010. His research interests include AI for Financial Security and AI Safety in Finance. He has authored more than 100 referred publications at prestigious international conferences and journals like The Innovation, IEEE TKDE, KDD, WWW, SIGIR, ACL, ICLR etc., and has served as Area Chair, SPC or PC members over top tier international conferences such as KDD, WWW, ACL, NeurIPS, ICML, AAAI, IJCAI, etc. 
\end{IEEEbiography}

\begin{IEEEbiography}[{\includegraphics[width=0.8in,height=1.25in,clip,keepaspectratio]{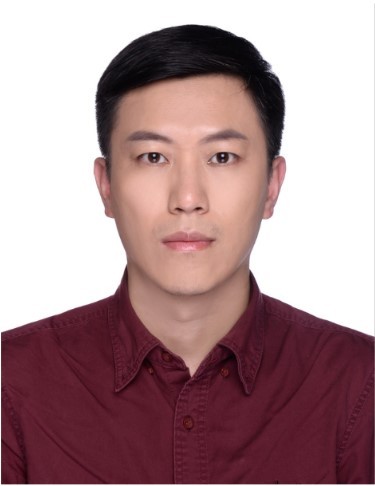}}]{Bin Wang}  received the Ph.D. degree from China National Digital Switching System Engineering and
Technological Research and Development Center.
He is a Professor with Zhejiang Key Laboratory of Artificial Intelligence of Things (AIoT) Network
and Data Security, Xidian University. His research
interests mainly include Internet of Things security,
cryptography, artificial intelligence security, and new
network security architecture.
\end{IEEEbiography}

\begin{IEEEbiography}[{\includegraphics[width=0.8in,height=1.25in,clip,keepaspectratio]{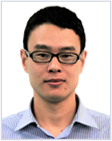}}]{Chenxu Wang} received the B.S. degree in communication engineering and the Ph.D. degree in control science and engineering from Xi’an Jiaotong University in 2009 and 2015, respectively. He is currently an Associate Professor with the School of Software Engineering, Xi’an Jiaotong University. His current research interests include graph data mining, complex network analysis, and network representation learning.
\end{IEEEbiography}

\begin{IEEEbiography}[{\includegraphics[width=0.8in,height=1.25in,clip,keepaspectratio]{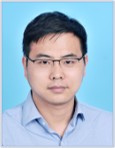}}]{Xiaobo Ma}  received the Ph.D. degree in control science and engineering from Xi’an Jiaotong University, Xi’an, China, in 2014. He was a Post-Doctoral Research Fellow with The Hong Kong Polytechnic University in 2015. He is currently a Professor with the MOE Key Laboratory for Intelligent Networks and Network Security, Faculty of Electronic and Information Engineering, Xi’an Jiaotong University. He is a Tang Scholar. His research interests include internet measurement and cyber security.
\end{IEEEbiography}

\begin{IEEEbiography}[{\includegraphics[width=0.9in,height=1.25in,clip,keepaspectratio]{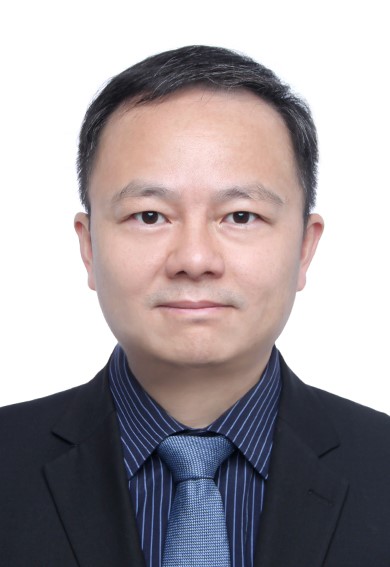}}]{Wei Wang} is a full Professor with Ministry of Education Key Lab for Intelligent Networks and Network Security, aka MOE KLINNS Lab, Xi'an Jiaotong University, China. He is also an adjunct Professor with school of computer science and technology, Beijing Jiaotong University. He received the Ph.D. degree from Xi'an Jiaotong University, in 2006. He was a Post-Doctoral Researcher with the University of Trento, Italy, from 2005 to 2006. He was a Post-Doctoral Researcher with TELECOM Bretagne and with INRIA, France, from 2007 to 2008. He was also a European ERCIM Fellow with the Norwegian University of Science and Technology (NTNU), Norway, and with the Interdisciplinary Centre for Security, Reliability, and Trust (SnT), University of Luxembourg, from 2009 to 2011. He was a faculty member with Beijing Jiaotong University from 2011 to 2024. His recent research interests lie in privacy-preserving computation and blockchain. He has authored or co-authored over 100 peer-reviewed articles in various journals and international conferences, including IEEE TDSC, IEEE TIFS, IEEE TSE, USENIX Security, ACM CCS, AAAI, Ubicomp, IEEE INFOCOM. He received the ACM CCS 2023 Distinguished Paper Award. He is an Elsevier ``highly cited Chinese Researchers''. He is an Associate Editor for IEEE Transactions on Dependable and Secure Computing, and an Editorial Board Member of Computers \& Security and of Frontiers of Computer Science. He is a vice chair of ACM SIGSAC China.
\end{IEEEbiography}

\vfill

\end{document}